\documentclass[11pt]{article}

\usepackage[preprint]{acl}

\usepackage{times}
\usepackage{latexsym}

\usepackage[T1]{fontenc}

\usepackage[utf8]{inputenc}

\usepackage{microtype}

\usepackage{inconsolata}

\usepackage{graphicx}

\usepackage{amssymb} 
\usepackage{amsmath}
\usepackage{booktabs}
\usepackage{multirow}
\usepackage[table]{xcolor}
\usepackage{subcaption} 
\usepackage{bm}
\usepackage{algorithm}
\usepackage{algorithmic}
\usepackage{amsthm} 
\usepackage{pifont}

\usepackage{tcolorbox}
\usepackage{colortbl}

\newtheorem{innercustomlemma}{Lemma}
\newenvironment{customlemma}[1]
  {\renewcommand\theinnercustomlemma{#1}\innercustomlemma}
  {\endinnercustomlemma}

\definecolor{def}{RGB}{132, 182, 248}
\newtcolorbox{defbox}[1][]{colback=def!5!white,colframe=def!60!black,boxsep=-4pt,grow to left by=4pt,left=10pt,grow to right by=4pt,right=10pt,top=10pt,bottom=10pt,#1}
\title{VecInfer: Efficient LLM Inference with Low-Bit KV Cache via Outlier-Suppressed Vector Quantization}

\author{
  Dingyu Yao$^{1,2*}$ , Chenxu Yang$^{1,2}$\ , Zhengyang Tong$^{1,2}$\ ,Zheng Lin$^{1,2}$\footnotemark[2] , Wei Liu$^{3}$,\\ \bf Jian Luan$^{3}$, 
   Weiping Wang$^{1}$\\
  $^{1}$Institute of Information Engineering, Chinese Academy of Sciences, Beijing, China\\
  $^{2}$School of Cyber Security, University of Chinese Academy of Sciences, Beijing, China\\
  $^{3}$MiLM Plus, Xiaomi Inc, Beijing, China\\
  \texttt{\{yaodingyu, yangchenxu, linzheng, wangweiping\}@iie.ac.cn}\\
  \texttt{\{liuwei40, luanjian\}@xiaomi.com}
}

\begin{document}
\maketitle
  \renewcommand{\thefootnote}{\fnsymbol{footnote}}
  \footnotetext[1]{\ \ Work done during an internship at Xiaomi Inc.}
  \footnotetext[2]{\ \ Corresponding Author: Zheng Lin.}
  \renewcommand{\thefootnote}{\arabic{footnote}}

\begin{abstract}
The Key-Value (KV) cache introduces substantial memory overhead during large language model (LLM) inference. Although existing vector quantization (VQ) methods reduce KV cache usage and provide flexible representational capacity across bit-widths, they suffer severe performance degradation at ultra-low bit-widths due to key cache outliers that hinder effective codebook utilization. To address this challenge, we propose VecInfer, a novel VQ method for aggressive KV cache compression while enabling efficient inference. By applying smooth and Hadamard transformations, VecInfer suppresses outliers in the key cache, enabling the codebook to comprehensively cover the original data distribution and thereby reducing quantization difficulty. To facilitate efficient deployment, we design an optimized CUDA kernel that fuses computation with dequantization to minimize memory access overhead. Extensive evaluations demonstrate that VecInfer consistently outperforms existing quantization baselines across both long-context understanding and mathematical reasoning tasks. With only 2-bit quantization, VecInfer achieves performance comparable to full precision, while delivering up to $\mathbf{2.7\times}$ speedup in large-batch self-attention computation and $\mathbf{8.3\times}$ reduction in single-batch end-to-end latency on Llama-3.1-8B with a 196k sequence length.
\end{abstract}
\section{Introduction}
Recent transformer-based large language models (LLMs)~\cite{comanici2025gemini25pushingfrontier,guo2025deepseek,yang2025qwen3technicalreport} have achieved remarkable success in long-context tasks, including multi-document understanding~\cite{bai-etal-2024-longbench} and complex reasoning~\cite{li202512surveyreasoning}.
To enable efficient inference, the Key-Value (KV) cache is a critical mechanism that stores previous key and value states to avoid redundant attention computations during autoregressive decoding. 
However, the KV cache size grows linearly with sequence length, posing significant challenges for efficient LLM inference and serving, particularly in terms of memory consumption and computational overhead.
\begin{table}[t!]
\centering
\setlength{\tabcolsep}{2.5pt}
\renewcommand{\arraystretch}{1.05}
\resizebox{\columnwidth}{!}{%
\begin{tabular}{|l|c|c|c|c||c|}
\hline
 &
  KIVI &
  ZipCache &
  CQ &
  MILLION & 
  VecInfer \\ \hline
Quantization Scheme &
  \cellcolor[HTML]{DFFFD3}SQ &
  \cellcolor[HTML]{DFFFD3}SQ &
  \cellcolor[HTML]{DFFFD3}VQ &
  \cellcolor[HTML]{DFFFD3}VQ &
  \cellcolor[HTML]{DFFFD3}VQ  \\ \hline
Bitwidth Flexibility &
  \cellcolor[HTML]{FFD3DF}$\downarrow$ &
  \cellcolor[HTML]{FFD3DF}$\downarrow$ &
  \cellcolor[HTML]{DFFFD3}$\uparrow$ &
  \cellcolor[HTML]{DFFFD3}$\uparrow$ &
  \cellcolor[HTML]{DFFFD3}$\pmb{\uparrow}$  \\ \hline
Accuracy @ Low-bit &
  \cellcolor[HTML]{DFFFD3}$\uparrow$ &
  \cellcolor[HTML]{DFFFD3}$\uparrow$ &
  \cellcolor[HTML]{FFD3DF}$\downarrow\downarrow$ &
  \cellcolor[HTML]{FFD3DF}$\downarrow\downarrow$ &
  \cellcolor[HTML]{DFFFD3}$\pmb{\uparrow}$  \\ \hline
Fused Attention &
  \cellcolor[HTML]{FFD3DF} \ding{55} &
  \cellcolor[HTML]{FFD3DF}\ding{55} &
  \cellcolor[HTML]{FFD3DF}\ding{55} &
  \cellcolor[HTML]{DFFFD3}\ding{51} &
  \cellcolor[HTML]{DFFFD3}\ding{51}  \\ \hline
Inference Speed &
  \cellcolor[HTML]{FFD3DF}$\downarrow\downarrow$ &
  \cellcolor[HTML]{FFD3DF}$\downarrow\downarrow$ &
  \cellcolor[HTML]{FFD3DF}$\downarrow\downarrow$ &
  \cellcolor[HTML]{DFFFD3}$\uparrow$ &
  \cellcolor[HTML]{DFFFD3}\pmb{$\uparrow$}  \\ \hline
\end{tabular}%
}

\caption{Comparison of different KV cache quantization methods across multiple dimensions. VecInfer achieves expected gains in both accuracy and efficiency.}
\label{tab: intro_table}

\end{table}

\begin{figure*}[t]
    \centering
    \begin{subfigure}[t]{0.49\textwidth}
        \centering
        \includegraphics[height=4.1cm, keepaspectratio]{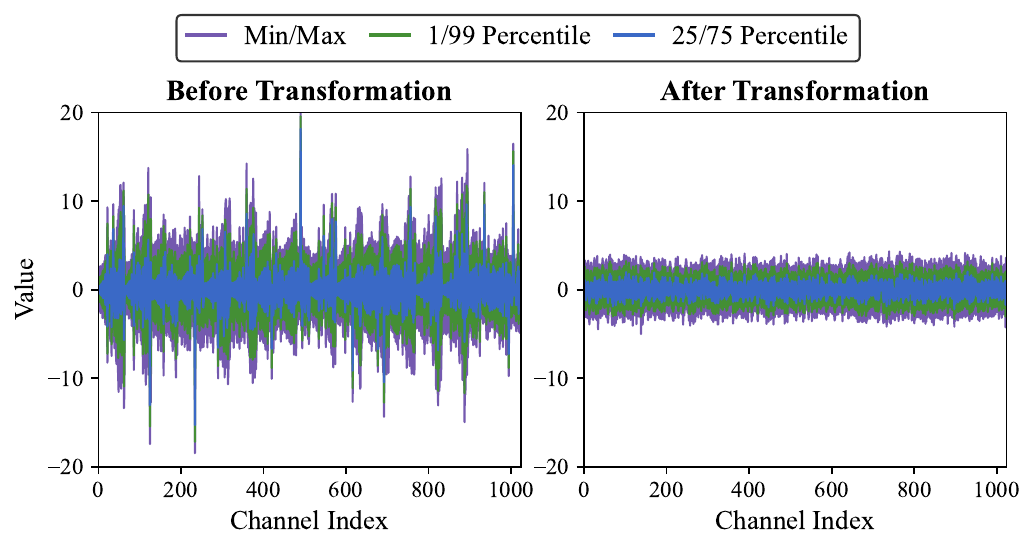}
        \caption{Channel-wise distribution of the key cache.}
        \label{fig: key_states}
    \end{subfigure}
    \hfill
    \begin{subfigure}[t]{0.49\textwidth}
        \centering
        \includegraphics[height=4.1cm, keepaspectratio]{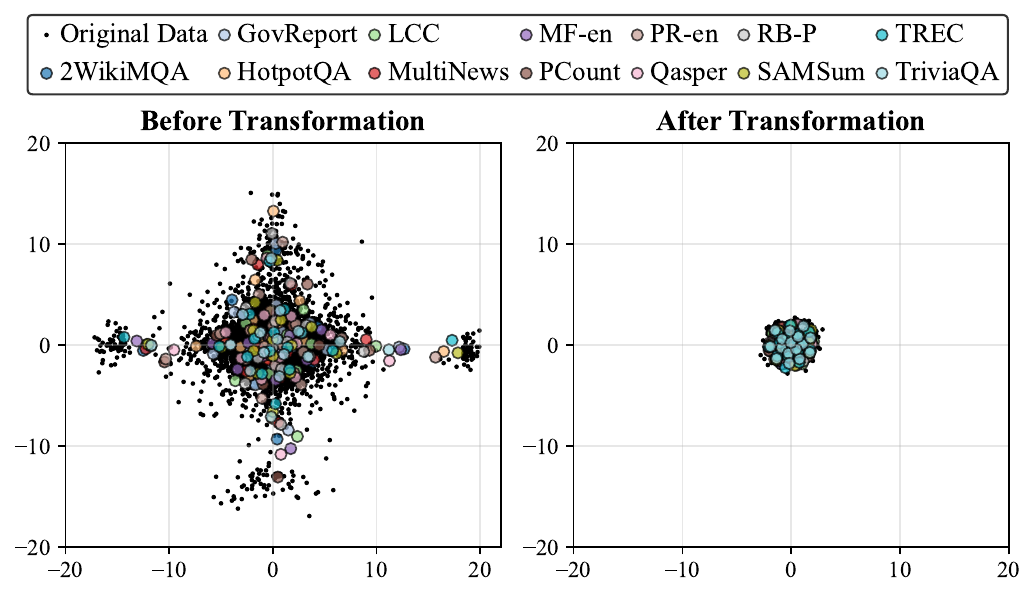}
        \caption{Codebook representation (\texttt{b2d4}) of the key cache.}
        \label{fig: codebook}
    \end{subfigure}
    \caption{Key cache distribution and codebook representation for Llama-3.1-8B-Instruct at layer 16. (a) Dual transformation reduces channel-wise variation and suppresses outliers, resulting in a more uniform distribution. (b) This uniformity \textbf{facilitates task-independent codebook representations} and \textbf{ensures comprehensive coverage of the original data distribution}.}
    \label{fig: intro2}
\end{figure*}

To reduce KV cache usage, quantization has emerged as a promising solution, primarily encompassing scalar quantization (SQ) and vector quantization (VQ).
SQ~\cite{hooper2024kvquant,liu2024kivi,he2024zipcache} maps floating-point values to fixed-point integers but offers limited flexibility across bit-widths.
In contrast, VQ~\cite{zhang2024kv,wang2025millionmasteringlongcontextllm,10946800} provides greater flexibility by mapping high-dimensional vectors to a finite set of codebook entries, where dequantization is reduced to an efficient lookup operation.
Table~\ref{tab: intro_table} summarizes representative KV cache quantization methods and compares their strengths and limitations from multiple perspectives.
Despite the memory savings, existing methods for low-bit KV cache still struggle to achieve the expected gains in both accuracy and efficiency.

To address these limitations, effective deployment of VQ-based KV cache must tackle two key challenges:
\textbf{\textit{(i) Lossless accuracy at low bit-widths}}: 
VQ typically quantizes KV cache along the token dimension for hardware compatibility, rendering it highly sensitive to outliers. As shown in Figure~\ref{fig: codebook}, outlier vectors lie far from any codebook centroids, and the learned centroids tend to be task-dependent, further increasing quantization difficulty.
\textbf{\textit{(ii) Hardware-aligned inference speedup}}:
Performing dequantization before attention computation introduces significant overhead, severely limiting practical speedup. 
Therefore, realizing actual speed gains requires hardware-friendly kernel designs that minimize memory access and optimize thread scheduling.

Building upon these analyses, we propose VecInfer, a novel VQ-based method for aggressive low-bit KV cache compression. 
Our approach first applies smooth and Hadamard transformations to the key cache, which reduces quantization difficulty while preserving the computational equivalence between queries and keys.
As demonstrated in Figure~\ref{fig: intro2}, this dual transformation suppresses outliers and produces a more uniform distribution, facilitating task-independent codebook representations and enabling the codebook to comprehensively cover the original data space.
Then we implement a fused dequantization-computation CUDA kernel featuring fine-grained tiled computation and asynchronous pipeline execution to enhance efficiency.

We evaluate VecInfer in terms of both accuracy and efficiency across diverse LLMs, focusing on long-context and mathematical reasoning tasks.
Experimental results show that VecInfer consistently outperforms existing baselines under lower bit-width configurations (1.25, 1.5, 2, 3, and 4 bits) and achieves substantial efficiency gains at both the kernel and end-to-end levels.
Specifically, for Llama-3.1-8B with 196k sequence length using 2-bit KV cache quantization, VecInfer achieves up to $2.7\times$ speedup on H100 and $2.8\times$ on A100 for large-batch self-attention computation compared to the FP16 counterpart,
and reduces single-batch end-to-end latency by $8.3\times$ on H100.
Below are the key contributions of our work:
\begin{itemize}
\item We identify outliers as a major challenge in VQ and propose \textbf{VecInfer}, a novel VQ-based method for KV cache that employs dual transformation to reduce quantization difficulty.
\item To enable efficient hardware acceleration, we design a fused dequantization-computation CUDA kernel with fine-grained tiled computation and asynchronous pipeline execution.
\item Extensive experiments show that VecInfer outperforms baselines across diverse quantization bit-widths, downstream tasks, and model architectures, while significantly reducing self-attention and end-to-end latency.
\end{itemize}

\section{Preliminaries}

\subsection{KV Cache and Attention}

The KV cache eliminates redundant attention computations by storing key-value states during LLM inference, which consists of prefilling and decoding phases.
During \textit{prefilling}, the prompt is processed to produce the first output token and initialize the KV cache with key-value states $\mathbf{K},\mathbf{V} \in \mathbb{R}^{N \times D}$, where $N$ is the number of prompt tokens and $D$ is the dim of attention.
During \textit{decoding}, the LLM performs autoregressive generation, producing the output sequence token by token.
For the current input states $\mathbf{q},\mathbf{k},\mathbf{v}\in \mathbb{R}^{1\times D}$, the KV cache is updated as $\mathbf{K}\!\leftarrow\!\mathrm{Concat}(\mathbf{K},\mathbf{k})$ and $\mathbf{V}\!\leftarrow\!\mathrm{Concat}(\mathbf{V},\mathbf{v})$.
The self-attention mechanism captures connections among all tokens in the context through the KV cache, computing the attention output as:
\begin{equation}
\mathbf{s}=\mathbf{qK}^\top/\sqrt{D},~\mathbf{p}=\mathrm{softmax}(\mathbf{s}), ~ \mathbf{o}=\mathbf{pV}.
\label{eq: attention}
\end{equation}
FlashAttention~\cite{dao2022flashattention,dao2024flashattention} is an IO-aware algorithm that reduces the memory overhead in the attention through tiling, re-computation, and online softmax operation.

\subsection{Vector Quantization}

\begin{figure}[t]
  \centering
  \includegraphics[width=1\linewidth]{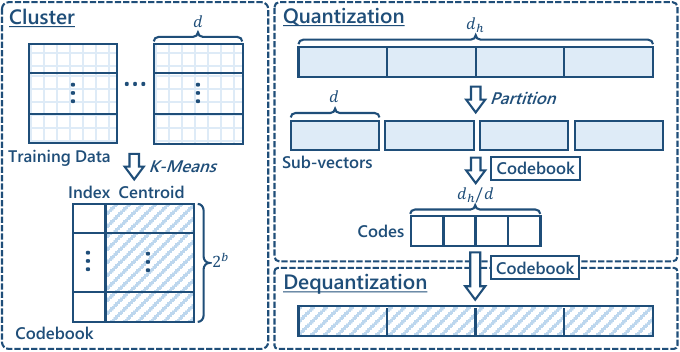} 
  \caption{Typical vector quantization pipeline.}
  \label{fig: vq}
\end{figure}

VQ~\cite{jegou2010product} maps continuous vector spaces to a finite set of representative codebook vectors, treating each vector as a quantization unit.
As shown in Figure~\ref{fig: vq}, VQ employs K-Means to construct a codebook $\mathcal{C}$ comprising $2^b$ centroids, each with $d$ dimensions. 
Given a $d_h$-dimensional vector $\mathbf{x} \in \mathbb{R}^{d_h}$, VQ partitions it into $d_h/d$ disjoint sub-vectors: $[\mathbf{x}_1,\ldots,\mathbf{x}_i,\ldots,\mathbf{x}_{d_h/d}]$. 
Each sub-vector $\mathbf{x}_i\in \mathbb{R}^{d}$ is then assigned the index of its nearest centroid in  $\mathcal{C}$, with the corresponding centroid index encoded as a $b$-bit representation:
\begin{equation}
    j^* = {\underset{j \in \{1,\dots,2^b\}}{\arg\min}} \|\mathbf{x}_i - \mathcal{C}_{j}\|^2, ~\mathrm{VQ}(\mathbf{x}_i,\mathcal{C})= j^*.
    \label{eq: vq}
\end{equation}
These sub-vector indices are then combined to form a compressed representation of the original vector $\mathbf{x}$. 
VQ significantly reduces memory usage.
Rather than storing the full $d_h$-dimensional vector using $d_h \times 2$ bytes (assuming 16-bit floating-point precision), VQ requires only $2^b \times d \times 2$ bytes for the codebook plus $(d_h/d) \times (b / 8)$ bytes for the indices.
However, during model inference, since the quantized data contains only codebook indices, direct arithmetic operations are not possible. Therefore, dequantization must be performed before each computation by retrieving the corresponding centroid from the codebook using the stored index.

\section{Methodology}

\subsection{Rethinking Quantization Difficulty}
\label{sec: difficulty}

While prior studies~\cite{wang2025millionmasteringlongcontextllm,zhang2024kv} show that VQ alleviates outlier issues compared to SQ, this advantage is limited in practice. 
As shown in Figure~\ref{fig: codebook} (left), outlier vectors remain distant from any codebook centroids, and these learned centroids exhibit highly task-dependence. Consequently, codebook entries are underutilized, which increases quantization difficulty.

\begin{figure}[t]
  \centering
  \includegraphics[width=1\linewidth]{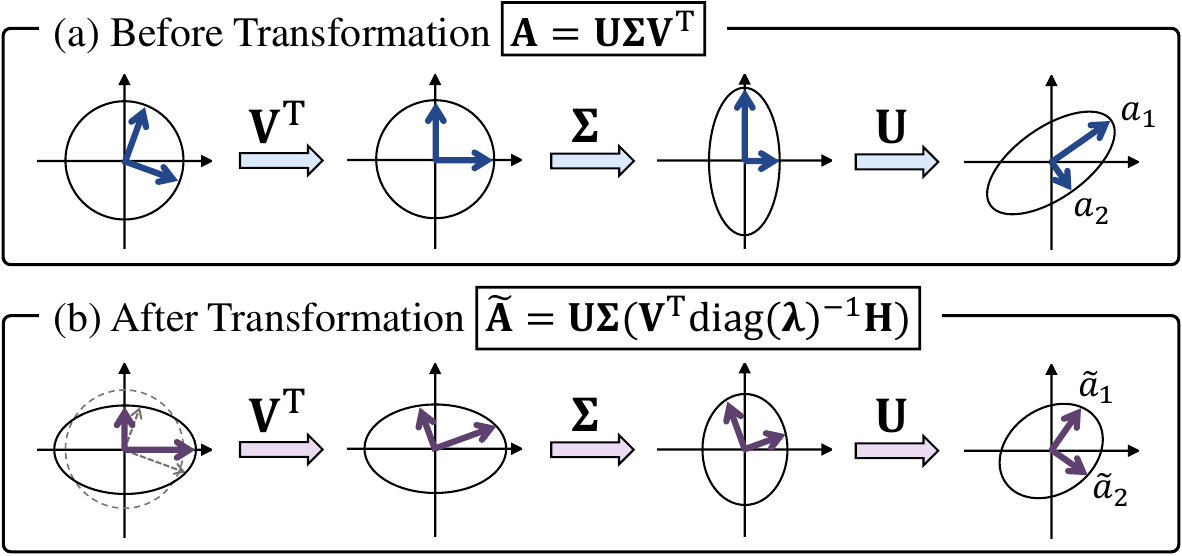} 
  \caption{Transformation from $\mathbf{V}^\top$ to $\mathbf{A}$ via SVD.}
  \label{fig: svd}
\end{figure}

Inspired by computational invariance in weight-activation transformations~\cite{xiao2023smoothquant,ashkboos2024quarot}, we study how smooth and Hadamard transformations reduce key cache quantization difficulty while ensuring computational invariance between queries and keys.
To analyze the effects of transformations, we employ singular value decomposition (SVD), which factorizes a matrix as $\mathbf{A}=\mathbf{U\Sigma V}^\top$, where the orthogonal matrices $\mathbf{U}$ and $\mathbf{V}$ are rotations, and the diagonal matrix $\mathbf{\Sigma}$ is stretch~\cite{lee2024infinigen}. 
Figure~\ref{fig: svd}(a) shows how the column vectors of $\mathbf{V}^\top$ are rotated and stretched to form the column vectors $a_1$ and $a_2$ of $\mathbf{A}$, representing the maximum and minimum values, respectively.
Figure~\ref{fig: svd}(b) shows that combining smooth and Hadamard transformations reduces the magnitude gap between $\tilde{a}_1$ and $\tilde{a}_2$, resulting in an outlier-free distribution.

As shown in Figure~\ref{fig: intro2}, the proposed dual transformation reduces channel-wise variation and suppresses outliers, producing a more uniform distribution. 
This uniformity facilitates task-independent codebook representations and ensures comprehensive coverage of the original data distribution.
Notably, applying these transformations individually leads to sub-optimal uniformity, with full details provided in Appendix~\ref{appendix: transformations}.

\subsection{Outlier-Suppressed Vector Quantization}
\begin{figure*}[t]
  \centering
  \includegraphics[width=1\linewidth]{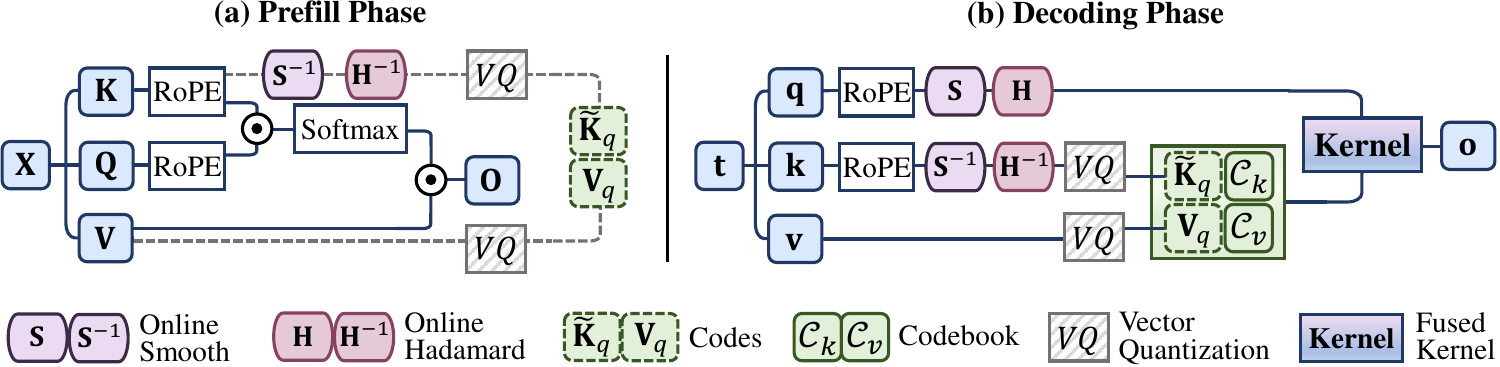} 
  \caption{Overview of VecInfer. During inference, dual transformation is applied before vector quantization.}
  \label{fig: method}
\end{figure*}
Building on the above analyses, we propose VecInfer, a VQ-based KV cache compression method that applies smooth and Hadamard transformations before quantization to suppress key cache outliers and reduce quantization difficulty.
The overall pipeline is illustrated in Figure~\ref{fig: method}.
\paragraph{Dual Equivalent Transformation.}
We first smooth the keys via channel-wise scaling using a factor $\bm{\lambda} \in \mathbb{R}^{D}$, and apply the inverse scaling to the queries to preserve computational invariance in the query–key multiplication:
\begin{equation}
\mathbf{q} \leftarrow \mathbf{q}\mathrm{diag}(\bm{\lambda}),~ \mathbf{K}  \leftarrow \mathbf{K}\mathrm{diag}(\bm{\lambda})^{-1}.
\end{equation}
Here, the scaling factor is pre-computed offline from calibration samples and defined as:
\begin{equation}
\bm{\lambda}_i=\sqrt{\mathrm{max}(\left| \mathbf{K}_i\right|)},~i=1,2,\ldots ,D,
\end{equation}
where $\mathbf {K}_i$ is the $i$-th channel of $\mathbf K$.

Since the smooth transformation reduces inter-channel variance without addressing intra-channel variance, significant percentile fluctuations persist (Figure~\ref{fig: key_states_appendix}(b)).
To further suppress outliers, we apply an orthogonal Hadamard matrix $\mathbf{H}_D$ satisfying $\mathbf{H}_D\mathbf{H}_D^\top=\mathbf{I}$.
For $D=2^k$, the Walsh–Hadamard matrix is defined recursively as:
\begin{equation}
\mathbf{H}_{2^k} = \frac{1}{\sqrt{2}}\begin{bmatrix}
\mathbf{H}_{2^{k-1}} & \mathbf{H}_{2^{k-1}} \\
\mathbf{H}_{2^{k-1}} & -\mathbf{H}_{2^{k-1}}
\end{bmatrix}, \mathbf{H}_1=\begin{bmatrix}
    1
\end{bmatrix}.
\label{eq: hadamard}
\end{equation}
By multiplying both queries and keys with $\mathbf{H}_D$, we ensure computational invariance between them:
\begin{equation}
\mathbf{q} \leftarrow \mathbf{q}\mathbf{H}_D,~ \mathbf{K}  \leftarrow \mathbf{K}\mathbf{H}_D.
\end{equation}
\begin{defbox}
\begin{customlemma}{1}[Hadamard]\label{lemma: Hadamard} 
For key states $\mathbf{K} \!\in \!\mathbb{R}^{N \times D}$ with $\mathrm{sign}(K_{i,j}) \overset{\text{i.i.d.}}{\sim} \mathrm{Uniform}\{-1, +1\}$, and a Hadamard matrix $\mathbf{H} \in \mathbb{R}^{D \times D}$ constructed as in Equation~\eqref{eq: hadamard}, the transformed matrix $\tilde{\mathbf{K}}=\mathbf{KH}$ exhibits approximately Gaussian distribution by the central limit theorem, thereby redistributing the outliers of $\mathbf{K}$.
\end{customlemma}
\end{defbox}
Lemma~\ref{lemma: Hadamard} suggests that the Hadamard rotation effectively redistributes outliers across neighboring elements, yielding a more uniform distribution and further reducing the difficulty of quantization.


In summary, after smooth and Hadamard transformations, the attention score can be rewritten as:
\begin{equation}
    \mathbf{s}
    =(\underbrace{\mathbf{q}\mathrm{diag}(\bm\lambda)\mathbf{H}_{D}}_{\tilde{\mathbf{q}}})\cdot(\underbrace{\mathbf{K}\mathrm{diag}(\bm{\lambda})^{-1}\mathbf{H}_{D}}_{\tilde{\mathbf{K}}})^{\top}.
\end{equation}

\paragraph{Vector-Quantized KV Cache in Attention.}
To seamlessly integrate VQ into attention, we pre-sample outlier-suppressed keys and pre-train the codebook via K-Means (see Figure~\ref{fig: vq}).
As shown in Figure~\ref{fig: method}, during prefilling, a dual transformation is applied to the keys, followed by VQ on the transformed keys $\tilde{\mathbf{K}}$ and original values $\mathbf{V}$:
\begin{equation}
\tilde{\mathbf{K}}_{q}=\mathrm{VQ}( \tilde{\mathbf{K}},\mathcal{C}_k),~\mathbf{V}_q=\mathrm{VQ}( \mathbf{V},\mathcal{C}_v),
\end{equation}
where $\mathrm{VQ}(\cdot)$ denotes the vector quantization function defined in Equation~\eqref{eq: vq}, and $\mathcal{C}_k, \mathcal{C}_v$ are the codebooks for keys and values, respectively.

During decoding, each newly arrived set of keys $\mathbf{k}$ undergoes online dual transformation. 
The transformed keys $\tilde{\mathbf{k}}$ and their corresponding values $\mathbf{v}$ are then quantized using the pre-trained codebooks.
The quantized results are subsequently concatenated with previously quantized pairs:
\begin{equation}
\begin{aligned}
\tilde{\mathbf{K}}_{q}&\leftarrow\mathrm{Concat}(\tilde{\mathbf{K}}_{q}, \mathrm{VQ}( \tilde{\mathbf{k}},\mathcal{C}_k)),\\
    \mathbf{V}_q&\leftarrow\mathrm{Concat}(\mathbf{V}_q,\mathrm{VQ}( \mathbf{v},\mathcal{C}_v)).
\end{aligned}
\end{equation}
For output consistency, we apply the inverse transformation to queries $\mathbf{q}$. Denote the dequantization operator by $\mathrm{VQ}^{-1}(\cdot)$. The attention computation is then given by:
\begin{equation}
\begin{aligned}
{\mathbf{s}}=\tilde{\mathbf{q}} (\mathrm{VQ}^{-1}&(\tilde{\mathbf{K}}_{q},\mathcal{C}_k))^{\top}/\sqrt{D},\\{\mathbf{p}}=\mathrm{softmax}({\mathbf{s}}),~&{\mathbf{o}}={\mathbf{p}}(\mathrm{VQ}^{-1}(\mathbf{V}_q,\mathcal{C}_v)).
\label{eq: attention2}
\end{aligned}
\end{equation}

Notably, even after transformations, keys exhibit higher quantization sensitivity than values (see Appendix~\ref{appendix: sensitivity} for details).
To preserve accuracy, higher bit-widths can be allocated to the keys.

\subsection{Hardware-efficient Customized Kernel}
\begin{figure*}[t]
  \centering
  \includegraphics[width=1\linewidth]{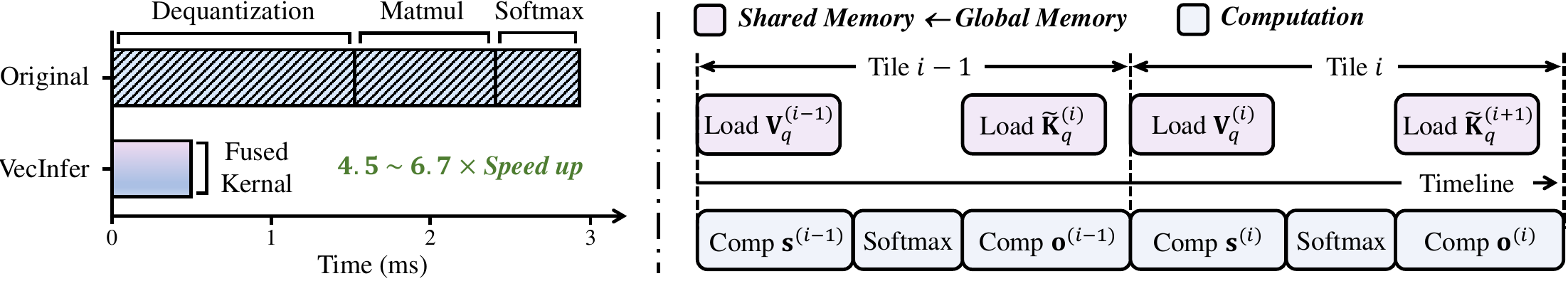} 
  \caption{\textbf{Left}: Attention kernel speed comparison between VecInfer and the non-fused baseline on H100. \textbf{Right}: Workflow of the VecInfer kernel with fine-grained tiled computation and asynchronous pipeline execution.}
  \label{fig: cuda}
\end{figure*}

\begin{figure}[t]
  \centering
  \includegraphics[width=1\linewidth]{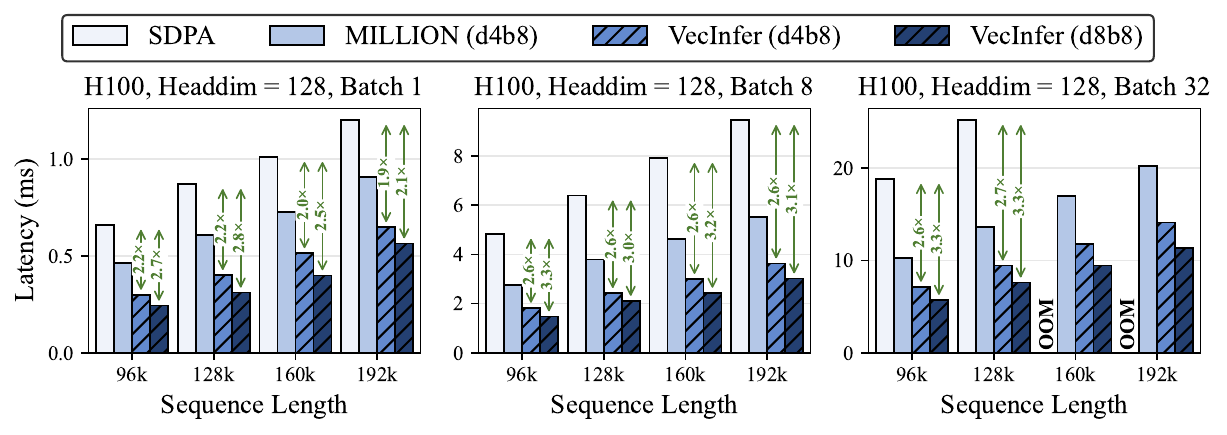} 
  \caption{Kernel performance on H100 (80GB). Additional results are provided in Figure~\ref{fig: kernal_all}.}
  \label{fig: kernal_latency}
\end{figure}

During autoregressive decoding, each newly generated token requires dequantization of the low-bit KV cache, which introduces substantial overhead and complexity.
To address this challenge, we propose a hardware-aligned kernel that fuses dequantization with attention computation. 
By minimizing global memory accesses, our kernel runs faster than the non-fused baseline (Figure~\ref{fig: kernal_latency} left).
Figure~\ref{fig: kernal_latency} (right) illustrates the VecInfer kernel workflow, and the architecture incorporates the following key optimizations:
\begin{enumerate}
\item \textbf{Fine-Grained Tiled Computation.}
Our implementation partitions the attention computation into tiles and loads them from global memory into shared memory, effectively mitigating the memory bandwidth bottleneck.
Specifically, we adopt a three-dimensional grid configuration of (\textit{batch\_size, num\_heads, num\_splits}), where each thread block contains 128 threads that collectively process a single tile of quantized key-value pairs to compute the corresponding partial attention output.
\item \textbf{Asynchronous Pipeline Execution.}
Our objective is to transfer quantized key-value pairs from global memory to shared memory for efficient access. 
To fully utilize CUDA cores, we leverage the \texttt{memcpy\_async} API to overlap memory transfers with computation.
During processing of the $i$-th tile, we asynchronously load value codes $\mathbf{V}^{(i)}_{q}$ while computing $\mathbf{s}^{(i)}$. 
Subsequently, during the computation of $\mathbf{o}^{(i)}$, we asynchronously prefetch key codes $\tilde{\mathbf{K}}_{q}^{(i+1)}$ for the next tile.
\end{enumerate}

Additionally, we optimize the shared memory layout of key and value codes to minimize bank conflicts and improve throughput.
The complete kernel algorithm is detailed in Appendix~\ref{appendix: kernel}.
As shown in Figure~\ref{fig: kernal_latency}, our optimized kernel achieves $2.6\sim3.3\times$ speedup over vanilla full attention on H100 for large-batch self-attention.

\section{Experiments}
\subsection{Experimental Setup}
\begin{table*}[t]
\setlength{\tabcolsep}{2.2pt}
\renewcommand{\arraystretch}{0.915}
\centering
\resizebox{\textwidth}{!}{
\begin{tabular}{lcc|ccccccccccccc|c}
\toprule

\multirow{3}{*}{\textbf{Method}} &\multirow{3}{*}{\textbf{Avg. bit}}  &\multirow{3}{*}{\textbf{Config}} & \multicolumn{2}{c}{\textbf{SD.QA}} & \multicolumn{2}{c}{\textbf{MD.QA}}& \multicolumn{2}{c}{\textbf{Sum.}}& \multicolumn{3}{c}{\textbf{FS.L}}& \multicolumn{2}{c}{\textbf{Code}} & \multicolumn{2}{c}{\textbf{Synth.}} &\multirow{3}{*}{\textbf{Avg.}} \\

\cmidrule(lr){4-5}\cmidrule(lr){6-7}\cmidrule(lr){8-9}\cmidrule(lr){10-12}\cmidrule(lr){13-14}\cmidrule(lr){15-16}
&&& Qspr & MulF & HQA & WMQA & GRpt & MulN & TREC & SMSM & TriQA & Repo & LCC & PsgC & PsgR \\

\midrule

\rowcolor[rgb]{0.957,0.957,0.957}
\textit{Llama-3.1-8B} & 16 & - & 45.9 & 53.8 & 55.2 & 46.6 & 34.6 & 27.5 & 72.5 & 43.8 & 91.6 & 56.4 & 63.2 & 8.0 & 99.5 & 53.7  \\
\midrule

KIVI & 3 & \texttt{b2g32} & 44.8 & 53.7 & 55.1 & 46.0 & 34.5 & 27.2 & 72.5 & 43.5 & 91.7 & 55.0 & 61.9 & 8.3 & 98.5 & 53.3\\

MILLION & 3 & \texttt{d4b12} & 44.1 & 50.3 & 46.6 & 38.6 & 33.4 & 27.0 & 69.5 & 41.0 & 91.2 & 52.5 & 55.9 & 3.1 & 87.5 & 49.3\\

\rowcolor[rgb]{0.87,0.94,1}
VecInfer & 3 & \texttt{d4b12} & 46.7 & 53.9 & 54.8 & 47.2 & 34.1 & 27.3 & 72.5 & 43.3 & 92.1 & 55.2 & 62.9 & 7.6 & 99.5 & \textbf{53.6}  \\

\midrule

KIVI & 2.25 & \texttt{b2g128} & 43.2 & 51.9 & 55.5 & 46.0 & 32.7 & 27.0 & 71.5 & 43.3 & 91.4 & 53.2 & 59.7 & 8.1 & 99.0 & 52.4\\

MILLION & 2 & \texttt{d4b8} & 39.4 & 49.3 & 52.5 & 40.0 & 26.8 & 24.6 & 69.0 & 38.3 & 91.7 & 44.9 & 50.7 & 7.8 & 91.5 & 48.2\\

\rowcolor[rgb]{0.87,0.94,1}
VecInfer & 2 & \texttt{d4b8} & 46.1 & 53.1 & 55.0 & 46.5 & 31.2 & 26.5 & 72.0   & 41.6 & 92.2 & 53.2 & 60.9 & 7.9 & 98.5 & \textbf{52.7}  \\
\midrule

KIVI & 1.5 & \texttt{b1g64} & 3.4 & 5.7 & 5.9 & 4.9 & 8.2 & 9.4 & 32.2 & 4.8 & 17.7 & 23.3 & 27.1 & 4.3 & 1.8 & 11.4 \\ 
MILLION & 1.5 & \texttt{d8b12} & 1.5 & 7.8 & 2.8 & 0.8 & 6.3 & 10.9 & 30.0 & 6.1 & 7.8 & 29.1 & 20.9 & 0.6 & 1.1  & 9.7\\
\rowcolor[rgb]{0.87,0.94,1}
VecInfer & 1.5 & \texttt{d8b12} & 43.7 & 52.2 & 54.7 & 46.2  & 29.2 & 26.1 & 71.0 & 39.4 & 91.7 & 51.4 & 60.2 & 7.6 & 99.5 & \textbf{51.8}\\
\rowcolor[rgb]{0.87,0.94,1}
VecInfer & 1.25 & \texttt{K-d8b12/V-d8b8} & 41.2 & 49.6 & 54.2 & 45.4 & 25.7 & 24.2 & 69.0 & 39.1 & 90.8 & 50.4 & 57.7 & 7.2 & 99.5 & \textbf{50.3}\\

\midrule

\rowcolor[rgb]{0.957,0.957,0.957}
\textit{Mistral-7B} & 16 & - & 38.6 & 49.7 & 51.0 & 36.4 & 34.3 & 26.5 & 76.0 & 47.6 & 88.5 & 60.6 & 59.4 & 6.0 & 96.5 & 51.5\\
\midrule

KIVI & 3 & \texttt{b2g32} & 37.3 & 48.3 & 50.3 & 36.6 & 34.1 & 26.3 & 76.0 & 47.2 & 88.4 & 59.1 & 57.6 & 8.0 & 92.0 & 50.8\\

MILLION & 3 & \texttt{d4b12}& 32.4 & 47.5 & 44.5 & 33.0 & 32.5 & 26.2 & 74.5 & 45.1 & 86.2 & 56.7 & 51.8 & 4.1 & 74.5 & 46.8 \\

\rowcolor[rgb]{0.87,0.94,1}
VecInfer & 3 & \texttt{d4b12} & 38.4 & 49.1 & 50.2 & 35.6 & 33.2 & 26.5 & 76.0   & 46.7 & 89.3 & 60.9 & 58.7 & 3.0    & 96.5 & \textbf{51.1} \\

\midrule

KIVI & 2.25 & \texttt{b2g128} & 36.4 & 47.8 & 49.0 & 30.9 & 32.3 & 26.4 & 76.0 & 46.6 & 89.2 & 56.9 & 56.2 & 4.6 & 87.0 & 49.2\\

MILLION & 2 & \texttt{d4b8} & 30.9 & 42.8 & 45.8 & 28.5 & 27.5 & 24.6 & 70.5 & 42.8 & 88.5 & 48.4 & 50.8 & 5.7 & 64.0 & 43.9\\

\rowcolor[rgb]{0.87,0.94,1}
VecInfer & 2 & \texttt{d4b8} & 36.7 & 49.1 & 51.7 & 34.0 & 31.3 & 26.5 & 76.0   & 45.4 & 89.4 & 60.5 & 57.7 & 3.1 & 92.5 & \textbf{50.3}\\
\midrule

KIVI & 1.5 & \texttt{b1g64} & 3.8 & 5.5 & 5.0 & 4.1 & 7.6 & 6.2 & 31.7 & 1.5 & 6.6 & 18.9 & 25.2 & 1.3 & 8.4 & 9.7\\ 
MILLION & 1.5 & \texttt{d8b12} & 8.5 & 18.2 & 11.2 & 8.2 & 15.7 & 20.9 & 54.0 & 15.8 & 27.1 & 26.7 & 27.0 & 2.1 & 3.8 & 18.4 \\
\rowcolor[rgb]{0.87,0.94,1}
VecInfer & 1.5 & \texttt{d8b12} & 33.2 & 45.4 & 48.2 & 32.1 & 27.2 & 25.2 & 73.0   & 44.4 & 88.2 & 58.3 & 56.0 & 5.0    & 87.5 & \textbf{48.0}\\
\rowcolor[rgb]{0.87,0.94,1}
VecInfer & 1.25 & \texttt{K-d8b12/V-d8b8} & 30.4 & 41.4 & 48.2 & 32.3 & 23.7 & 23.8 & 69.0   & 41.9  & 88.0 & 55.8 & 54.9  & 3.7 & 90.5 & \textbf{46.5}\\

\midrule

\rowcolor[rgb]{0.957,0.957,0.957}
\textit{Qwen2.5-14B} & 16 & - & 45.4 & 53.9 & 61.6 & 57.9 & 29.7 & 21.9 & 76.5 & 47.7 & 90.1 & 48.8 & 61.3 & 9.0 & 98.6 &54.0 \\

\midrule

KIVI & 3 & \texttt{b2g32} & 46.4 & 51.5 & 61.5 & 58.4 & 28.8 & 21.5 & 77.0 & 47.2 & 89.7 & 48.5 & 61.4 & 10.4 & 92.0 &53.4\\

MILLION & 3 & \texttt{d4b12}& 31.9 & 50.0 & 56.2 & 47.4 & 27.2 & 21.0 & 76.0 & 46.7 & 90.5 & 40.1 & 45.9 & 2.4 & 72.5 & 46.8 \\

\rowcolor[rgb]{0.87,0.94,1}
VecInfer & 3 & \texttt{d4b12} & 46.1 & 52.2 & 61.2 & 58.0 & 29.0 & 21.8 & 77.0   & 47.3  & 89.7 & 49.1 & 61.8 & 9.6 & 97.0 & \textbf{53.9} \\

\midrule

KIVI & 2.25 & \texttt{b2g128} & 41.6 & 49.2 & 61.8 & 58.2 & 27.4 & 21.4 & 76.5 & 46.9 & 90.2 & 46.7 & 60.0 & 11.7 & 74.6 & 51.2\\

MILLION & 2 & \texttt{d4b8} & 24.8 & 45.0 & 57.2 & 48.3 & 23.1 & 20.7 & 72.5 & 41.5 & 86.4 & 37.2 & 40.8 & 15.0 & 58.5 & 43.9 \\

\rowcolor[rgb]{0.87,0.94,1}
VecInfer & 2 & \texttt{d4b8} & 43.7 & 51.3 & 59.6 & 57.4 & 26.8 & 21.3 & 77.0   & 46.2  & 90.1 &   46.7 & 60.7 & 13.0& 90.8 & \textbf{52.7} \\
\midrule

KIVI & 1.5 & \texttt{b1g64} & 3.4 & 4.1 & 3.5 & 3.1 & 10.1 & 7.9 & 30.9 & 5.5 & 11.6 & 23.3 & 23.4 & 1.8 & 2.8 & 10.1\\ 
MILLION & 1.5 & \texttt{d8b12} & 3.9 & 3.9 & 0.9 & 0.9 & 7.2 & 8.4 & 42.0 & 11.6 & 5.3 & 28.6 & 21.3 & 0.5 & 0.0 & 10.3 \\
\rowcolor[rgb]{0.87,0.94,1}
VecInfer & 1.5 & \texttt{d8b12} & 38.2 & 45.3 & 58.2 & 54.9 & 24.0 & 20.5 & 72.5 & 42.8 & 89.4 & 43.6 & 57.5 & 9.0 & 88.5 & \textbf{49.6}\\
\rowcolor[rgb]{0.87,0.94,1}
VecInfer & 1.25 & \texttt{K-d8b12/V-d8b8} & 33.8 & 42.7 & 58.1 & 50.5 & 20.5 & 18.4 & 68.0   & 40.3 & 88.3 & 42.0 & 54.4 & 9.5 & 84.8 & \textbf{47.0}\\

\bottomrule
 
\end{tabular}}

\caption{The evaluation accuracy results on LongBench under different quantization configurations.  }
\label{tab: longbench}
\end{table*}

\paragraph{Models and Tasks.}

In this paper, we conduct experiments across a diverse range of LLMs, including Llama-3.1-8B-Instruct, Mistral-7B-Instruct-v0.3, Qwen2.5-14B-Instruct, DeepSeek-R1-Distill-Llama-8B, and DeepSeek-R1-Distill-Qwen-14B, Qwen3-8B.
To assess long-context performance, we evaluate VecInfer on 13 tasks from LongBench~\cite{bai-etal-2024-longbench}, which spans six categories: single/multi-document question answering, summarization, few-shot learning, code completion, and synthetic tasks.
To evaluate reasoning ability, we use three datasets: GSM8K \cite{cobbe2021trainingverifierssolvemath}, MATH500 \cite{hendrycks2021measuring}, AIME24, and AMC2023. We follow the recommended sampling parameters, setting the temperature to 0.6 and the top-$p$ to 0.95. The evaluation metric employed is Pass@1 accuracy. For GSM8K, MATH500, and AMC2023, the maximum output length is set to 16,384 tokens, while for AIME24, the maximum output length is set to 32,768 tokens.
Refer to Appendix~\ref{appendix: benchmark} for further details.

\begin{table*}[t]
\centering
\small
\setlength{\tabcolsep}{1.1pt}
\renewcommand{\arraystretch}{0.95}
\resizebox{\textwidth}{!}{
\begin{tabular}{lcc|cccc|cccc|cccc}
\toprule
\multirow{3}{*}{\textbf{Method}} & \multirow{3}{*}{\textbf{Avg. bit}} & \multirow{3}{*}{\textbf{Config}} & \multicolumn{4}{c}{\textbf{DS-R1-Distill-Llama-8B}} &  \multicolumn{4}{c}{\textbf{DS-R1-Distill-Qwen-14B}} &  \multicolumn{4}{c}{\textbf{Qwen3-8B}} \\ 

\cmidrule(lr){4-7}\cmidrule(lr){8-11}\cmidrule(lr){12-15}
&&& {\scriptsize \textbf{MATH500}} 
   & {\scriptsize \textbf{GSM8K}} 
   & {\scriptsize \textbf{AIME24}} 
   & {\scriptsize \textbf{AMC}} 
   & {\scriptsize \textbf{MATH500}} 
   & {\scriptsize \textbf{GSM8K}} 
   & {\scriptsize \textbf{AIME24}}
   & {\scriptsize \textbf{AMC}}
   & {\scriptsize \textbf{MATH500}} 
   & {\scriptsize \textbf{GSM8K}} 
   & {\scriptsize \textbf{AIME24}}
   & {\scriptsize \textbf{AMC}}\\

\midrule
\rowcolor[rgb]{0.957,0.957,0.957}
Baseline & 16 & -  & 86.6 & 90.4 & 47.5 & 86.8 & 92.6 & 95.7 & 66.2 & 93.1 & 94.0 & 96.0 & 72.9 & 90.0
\\
\midrule
KIVI & 4.25 &\texttt{b4g128}  & 87.8 & 90.1 & 45.8 & 86.8 & 93.6 & 95.4 & \textbf{67.1} & 92.8 & 93.8
& 96.1 & 72.2 & 90.0

\\
MILLION & 4 & \texttt{d2b8} & 86.8 &89.7& 40.4 & 85.6& 92.6 & 94.6 & 55.0 & 90.9 & 46.9 & 76.2 & 8.9&  24.3 \\
\rowcolor[rgb]{0.87,0.94,1}
VecInfer & 4 & \texttt{d2b8} & \textbf{88.2} & \textbf{90.9} & \textbf{46.2} & \textbf{87.2} & \textbf{93.6} &\textbf{95.5} & 66.7& \textbf{92.9}  & \textbf{94.0} & \textbf{96.1} & \textbf{73.9} & \textbf{90.6}\\
\midrule
KIVI & 3 &\texttt{b2g32} & 86.8 & 88.2 & 37.1 & 81.8 & 92.0 & \textbf{95.7} & 62.0 & 90.3 & 93.4 & 94.5 & 72.5 & 86.8 

  \\
MILLION & 3 & \texttt{d4b12} & 86.0 & 89.0 & 35.0 & 79.6&91.4 & 94.7  & 47.1 & 87.8 & 85.4 & 94.0 & 12.7 & 61.5   \\
\rowcolor[rgb]{0.87,0.94,1}
VecInfer & 3 & \texttt{d4b12} & \textbf{88.2} & \textbf{90.1} &\textbf{44.2} & \textbf{88.2} & \textbf{94.2} & 95.0 & \textbf{67.1} & \textbf{91.3} & \textbf{93.4} & \textbf{95.3} &\textbf{75.0} & \textbf{90.0} \\

\midrule
KIVI & 2.25 &\texttt{b2g128} & 74.6  & 84.9 & 16.3 & 65.9 & 91.2 & 94.5 & 47.5 & 81.8 & 88.4 & 93.1 & 57.4 & 84.0

 \\
MILLION & 2 & \texttt{d4b8} & 16.8 & 31.4& 0.0& 5.4& 38.8 & 60.9 & 0.0 &17.6  & 11.3 & 11.2& 0.0 & 12.5 \\
\rowcolor[rgb]{0.87,0.94,1}
VecInfer & 2 &  \texttt{K-d4b10}/\texttt{V-d8b12}  & \textbf{80.0} & \textbf{87.0} & \textbf{26.7} & \textbf{78.5} & \textbf{92.6} &\textbf{94.7} & \textbf{53.8} & \textbf{86.3} & \textbf{90.6} & \textbf{95.6} &  \textbf{67.1} & \textbf{86.3}\\

\bottomrule
\end{tabular}
    }
  \caption{Performance of large reasoning models on mathematical reasoning tasks. For AIME24 and AMC2023, 8 completions are generated per question.} 
  \label{tab: math}
\end{table*}

\paragraph{Baselines.}

We evaluate VecInfer against two representative baselines: KIVI~\cite{liu2024kivi} (scalar quantization) and MILLION~\cite{wang2025millionmasteringlongcontextllm} (vector quantization).
We adopt the notation \texttt{b$n$g$m$} for KIVI, where the KV cache is quantized to $n$-bit precision with a group size of $m$.
Both MILLION and VecInfer settings follow the notation \texttt{d$n$b$m$}, where each sub-vector has dimension $n$ and codes are encoded using $m$ bits.
Additionally, the residual length for all methods is set to 128.

\paragraph{Implementation Details.}

The smoothing factors are calibrated offline using 256 random samples from the Pile dataset~\cite{gao2020pile}, each consisting of 512 tokens.
This calibration process is efficient, requiring only a few seconds on an H100 GPU.
The codebook is pre-trained on the Qasper dataset using K-means clustering, with the maximum number of iterations set to 30.
Appendix~\ref{appendix: codebook} demonstrates that the codebook trained by our method is task-independent.

\subsection{Accuracy Evaluation}

\paragraph{Long Context Tasks.}
To evaluate VecInfer's long-context performance, we conduct experiments on 13 datasets from LongBench, quantizing the KV cache under different quantization configurations.
As shown in Table~\ref{tab: longbench}, VecInfer consistently outperforms baselines across the average precision range of 1.25 to 4 bits, thereby demonstrating the effectiveness of the proposed dual transformation.
Furthermore, when using 2-bit precision for KV cache storage, VecInfer demonstrates only a 2.1\% average accuracy drop, while achieving an 14.5\% average performance improvement compared to MILLION, another VQ-based method.

\paragraph{Complex Reasoning Tasks.}

We evaluate the performance of long-CoT LLMs on mathematical reasoning tasks~\cite{yang2025testtimepromptintervention,yang2025dynamicearlyexitreasoning}, as shown in Table~\ref{tab: math}.
When precision is reduced to 2-bit, both KIVI and MILLION experience significant performance degradation, failing to generate coherent responses.
In contrast, VecInfer shows minimal performance degradation on complex reasoning tasks.
We also find that model type and task difficulty significantly influence performance degradation.
Specifically, at equivalent compression ratios, DeepSeek-R1-Distill-Qwen-14B and Qwen3-8B exhibit superior quantization tolerance compared to DeepSeek-R1-Distill-Llama-8B, while complex tasks (e.g., AIME24) suffer more substantial performance drops than simpler ones (e.g., GSM8K).

\subsection{Efficiency Evaluation}
\paragraph{End-to-End Latency.}

\begin{figure}[t]
  \centering
  \includegraphics[width=1\linewidth]{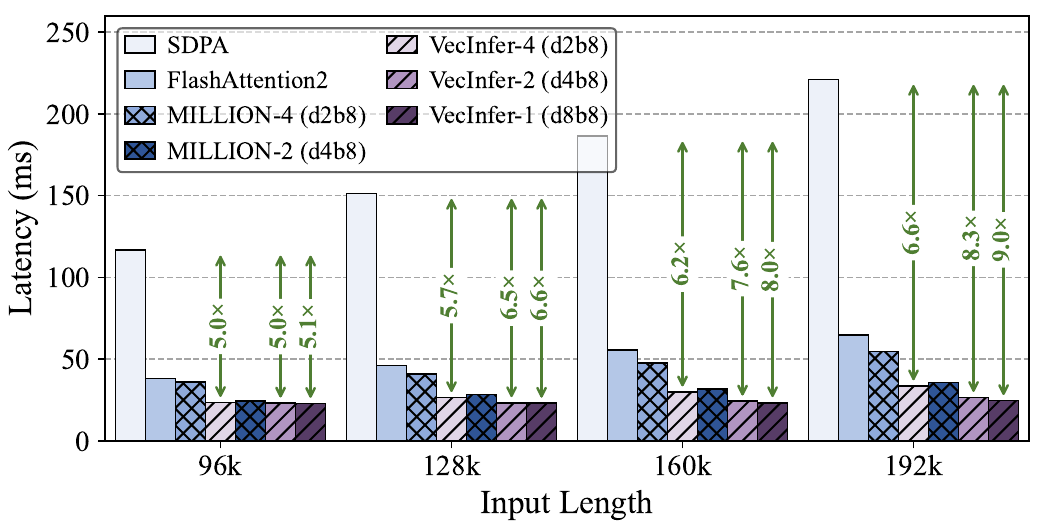} 
  \caption{Decoding latency comparison between VecInfer and baselines for Llama-3.1-8B-Instruct on an H100 GPU. Notably, KIVI runs into OOM.}
  \label{fig: latency}
\end{figure}

We evaluate VecInfer's end-to-end latency and compare it with several baselines, 
including Scaled Dot-Product Attention (SDPA), FlashAttention2~\cite{dao2024flashattention}, and MILLION. 
At a 64k sequence length, KIVI~\cite{liu2024kivi} runs into out-of-memory (OOM) errors due to missing fused kernel support.
As shown in Figure~\ref{fig: latency}, VecInfer consistently achieves lower end-to-end latency than previous methods.
For instance, at an input length of $l_{input}=192\text{k}$ and an output length of $l_{output}=129$, VecInfer achieves decoding speedups of $9.0\times$, $8.3\times$, and $6.6\times$ under 1-bit, 2-bit, and 4-bit configurations, respectively.
Moreover, the speedup advantage grows with sequence length.

\begin{figure}[t]
  \centering
  \includegraphics[width=1\linewidth]{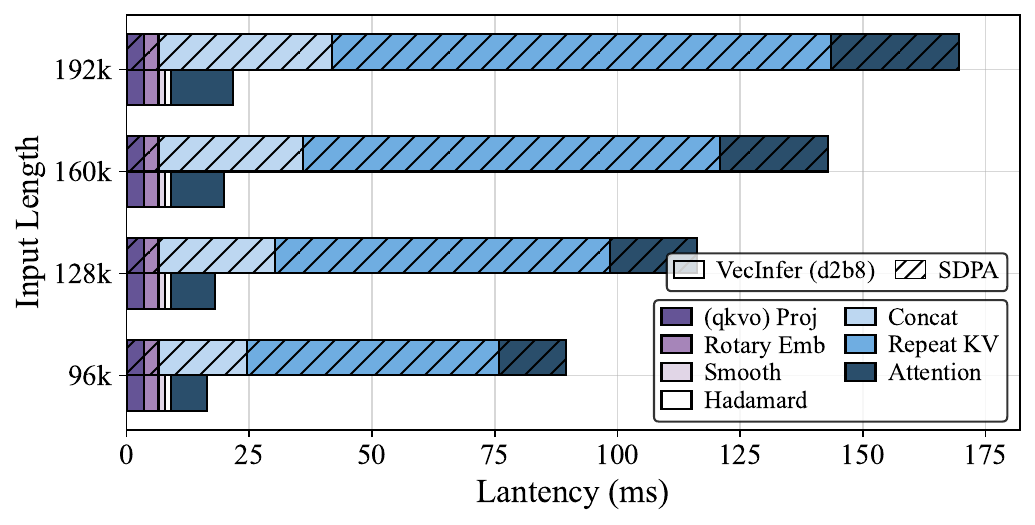} 
  \caption{Latency breakdown of attention blocks for Llama-3.1-8B-Instruct on an H100 GPU.}
  \label{fig: breakdown}
\end{figure}

\paragraph{Latency Breakdown.}
Figure~\ref{fig: breakdown} shows the latency breakdown of attention blocks across varying input lengths. Relative to SDPA, VecInfer reduces global memory read/write overhead by eliminating costly concatenation and repetition, thereby improving efficiency. 
At a sequence length of 196k under the 2-bit configuration, VecInfer achieves a $2.0\times$ speedup in self-attention, mainly by fusing all attention operations and dequantization into a single GPU kernel.
The additional cost of smooth and Hadamard transformations is negligible, exerting minimal impact on overall performance.

\subsection{Ablation Study}

\paragraph{Effects of Different Transformations.}

Table~\ref{tab: ab1} reports the comparative performance of different transformations on LongBench. Starting from VQ-only as the baseline, the smooth and Hadamard transformations independently improve average performance by 4.9\% and 14.1\%, respectively.
Importantly, their combination delivers substantially greater gains than either transformation alone.
The sequential application of smooth-then-Hadamard and Hadamard-then-smooth transformations yields comparable performance improvements.
\begin{table}[t]
\centering
\small
\setlength{\tabcolsep}{2.5pt}
\renewcommand{\arraystretch}{0.95}
\resizebox{\linewidth}{!}{
\begin{tabular}{l|cccccc|c}
\toprule

\textbf{Method} & \textbf{SD.QA} & \textbf{MD.QA} & \textbf{Sum.} & \textbf{FS.L} & \textbf{Code} & \textbf{Synth.} & \textbf{Avg.}\\
\midrule

\multicolumn{8}{c}{\textit{Llama-3.1-8B-Instruct}}\\
\midrule

\textit{Original}& 38.2 & 47.2 & 23.6 & 57.5 & 51.7 & 52.6 & 46.1\\
$\mathbf{S}$ & 42.6  & 48.4  & 25.0  & 63.6 & 51.7 & 50.8 & 48.3\\

$\mathbf{H}$ & 47.2  & 49.6 & 26.1  & 66.7 & 55.2 & 53.6 & 51.0\\

$\mathbf{H+S}$ & 46.8 & \textbf{50.6} & 26.7 & 67.3 & 55.1 & \textbf{53.8} & 51.4\\

$\mathbf{S+H}$ & \textbf{47.9}&  50.5 & \textbf{27.7} & \textbf{67.4} & \textbf{55.8} & 53.6 & \textbf{51.8}\\

 \midrule
\multicolumn{8}{c}{\textit{Qwen2.5-14B-Instruct}}\\
\midrule
\textit{Original} & 24.9 & 50.0 & 18.7 & 57.4 & 46.1 & 44.7 & 41.6\\
$\mathbf{S}$ & 32.6 & 52.6 & 19.7 & 58.9 & 48.4 &42.0 & 43.7 \\

$\mathbf{H}$ & 41.6 & 56.0 & 21.2 & 67.6 & 48.9 & 48.6 & 49.0\\

$\mathbf{H+S}$ & \textbf{43.3} & \textbf{56.8} & 21.4 & 67.9 & 49.2 & 48.6 & 49.4\\

$\mathbf{S+H}$ &41.8 & 56.6&  \textbf{22.3} & \textbf{68.3} & \textbf{50.6} & \textbf{48.9} & \textbf{49.6} \\

\bottomrule
\end{tabular}
}
\caption{Ablation study on the effect of different transformations under 1.5-bit, where $\mathbf{S}$ and $\mathbf{H}$ represent smooth and Hadamard transformations, respectively.}

  \label{tab: ab1}
\end{table}
\paragraph{Effects of Codebook Size.}

Codebook size directly influences the representational capacity of VQ and thus affects performance.
As previously discussed, codebook size is calculated as $2^b \times d \times 2$ bytes.
Table~\ref{tab: ab2} indicates that for a given bit-width, increasing codebook size consistently enhances accuracy.
However, this improvement comes at the cost of elevated shared memory overhead, which degrades computational efficiency. 
To balance this accuracy-efficiency trade-off, we adopt codebook sizes of $ 2^{8} \times 4 \times 2$ bytes for 2-bit quantization and $ 2^{12} \times 8 \times 2$ bytes for 1.5-bit quantization.
\begin{table}[t]
\centering
\small
\setlength{\tabcolsep}{1.6pt}
\renewcommand{\arraystretch}{0.94}
\resizebox{\linewidth}{!}{
\begin{tabular}{l|c|cccccc|c}
\toprule

\textbf{Avg. bit} & \textbf{Config} & \textbf{SD.QA} & \textbf{MD.QA} & \textbf{Sum.} & \textbf{FS.L} & \textbf{Code} & \textbf{Synth.} & \textbf{Avg.}\\
\midrule
\multicolumn{9}{c}{\textit{Llama-3.1-8B-Instruct}}\\
\midrule
\multirow{3}{*}{2} & \texttt{d8b16} &48.9 & 50.6 & 29.3 &68.9 & 58.0 & 53.5 & 52.8\\
                       & \texttt{d4b8}  &49.6 & 50.8 & 28.8 & 68.6 & 57.0 & 53.2 & 52.7 \\
                       & \texttt{d2b4}  & 48.8& 50.4 &  29.5 & 68.8& 56.3 & 53.6 & 52.6\\
\midrule
\multirow{3}{*}{1.5} & \texttt{d8b12} & 47.9&  50.5 & 27.7 & 67.4 & 55.8 & 53.6 & 51.8\\
                         & \texttt{d4b6}  & 46.2 & 50.4 & 22.8 & 67.0 & 53.8 & 53.3 & 50.2 \\
                         & \texttt{d2b3}  & 43.7 & 49.6&  20.1 & 66.7 &51.7 & 53.8 & 49.1 \\
\midrule
\multicolumn{9}{c}{\textit{Mistral-7B-Instruct-v0.3}}\\
\midrule
\multirow{3}{*}{2} & \texttt{d8b16} & 43.2& 42.4& 28.8 & 70.4 & 59.6& 49.0 & 50.5 \\
                       & \texttt{d4b8}  & 42.9& 42.9 & 28.9 & 70.3  & 59.1 & 47.9  & 50.3 \\
                       & \texttt{d2b4}  & 42.5 & 42.3 & 28.1&69.7 & 58.6& 46.8 &  49.7\\
\midrule
\multirow{3}{*}{1.5} & \texttt{d8b12} &39.3 & 40.2&  26.2 & 68.6 & 57.1 & 46.3 & 48.0 \\
                         & \texttt{d4b6}  & 38.2 & 38.4 & 24.4& 67.4 &56.6 & 47.8 & 47.1\\
                         & \texttt{d2b3}  & 37.1& 37.6&22.5 &64.1 &52.7 &47.6 & 45.1 \\
\bottomrule
\end{tabular}
}
\caption{Ablation study on the effect of codebook size.}
  \label{tab: ab2}
\end{table}

\section{Related Works}

\paragraph{KV Cache Quantization.}
Existing methods optimize KV cache memory usage through quantization while preserving performance. These methods are typically categorized into SQ and VQ. 
SQ compresses data by encoding floating-point values as low-bit integers.
KIVI~\cite{liu2024kivi} reduces quantization errors using per-channel quantization for key and per-token quantization for value.
MiKV~\cite{yang2024no}, ZipCache~\cite{he2024zipcache}, and RotateKV~\cite{su2025rotatekvaccuraterobust2bit} employ mixed-precision per-token quantization, preserving high precision for salient tokens.
TailorKV~\cite{yao-etal-2025-tailorkv} identifies that different layers exhibit varying compression preferences and quantizes select quantization-friendly layers.
In contrast, VQ encodes high-dimensional vectors using a finite codebook, improving bit utilization by leveraging inter-element correlations.
CQ~\cite{zhang2024kv} and MILLION~\cite{wang2025millionmasteringlongcontextllm} group multiple channels together for quantization by utilizing cross-channel dependencies. VQ-LLM~\cite{10946800} adaptively stores different codebook entries across the GPU's memory hierarchy.
However, a key limitation of these methods is that outlier vectors deviate significantly from cluster centroids, substantially impairing quantization accuracy.

\paragraph{Efficient Attention.}
Beyond quantization, several studies have explored sparse attention mechanisms to enhance efficiency in LLMs. 
Eviction-based methods, including StreamingLLM~\cite{xiao2024efficient}, H2O~\cite{zhang2023h2o}, and SnapKV~\cite{li2024snapkv}, selectively retain essential key-value pairs while permanently discarding less critical ones.
Selection-based methods such as NSA~\cite{yuan-etal-2025-native}, Quest~\cite{tang2024quest}, and MoBA~\cite{lu2025mobamixtureblockattention} identify the most important token blocks and optimize sparse attention patterns at the block level to enable efficient contiguous memory access.
Importantly, these sparse attention techniques are orthogonal to quantization and can be effectively combined to achieve significant reductions in both memory footprint and computational costs during inference.
Additionally, FlashAttention~\cite{dao2022flashattention,dao2024flashattention} employs a tiling strategy that partitions attention computation into blocks and performs operations within shared memory, thus reducing global memory access overhead.
\section{Conclusion}

In this paper, we propose VecInfer, a novel VQ method for aggressive KV cache compression while enabling efficient inference.
VecInfer employs smooth and Hadamard transformations to suppress outliers in the key cache and improve codebook utilization, thereby reducing quantization difficulty.
Extensive experiments show that VecInfer outperforms baselines on long-context and mathematical reasoning tasks.
Moreover, by fusing computation and dequantization into a single CUDA kernel, VecInfer significantly reduces attention and end-to-end latency.
VecInfer facilitates the deployment of LLMs on resource-constrained GPUs, thereby extending their applicability while maintaining both accuracy and efficiency.
\section*{Limitations}
The methodology offers multiple directions for further enhancement.
First, while combining vector quantization with sparse attention patterns for mixed-precision KV cache compression is a promising approach, the trade-offs between accuracy and efficiency remain to be thoroughly explored. 
Second, integrating VecInfer seamlessly into existing serving frameworks (e.g., vLLM~\cite{kwon2023efficientmemorymanagementlarge} and SGLang~\cite{zheng2024sglangefficientexecutionstructured}) poses practical challenges, particularly because many frameworks lack native support or flexible APIs for KV cache compression, which can complicate deployment.

\bibliography{custom}

\appendix
\section{Related Works}
\subsection{FlashAttention}
The query, key, and value matrices $\mathbf{Q}$, $\mathbf{K}$, and $\mathbf{V}$ are defined with dimensions $N \times D$, where $N$ denotes the sequence length and $D$ represents the dimension of attention.
The self-attention computation is formulated as follows:
\begin{equation}
\mathbf{S}\! = \!\mathbf{Q}\mathbf{K}^{\!\top} / \sqrt{D},~ 
\mathbf{P} \!= \!\mathrm{softmax}(\mathbf{S}),~
\mathbf{O} \!=\! \mathbf{P}\mathbf{V}.
\end{equation}
Standard attention implementation involves computing large intermediate matrices, specifically the $N \times N$ matrices $(\mathbf{S}$, $\mathbf{P})$, which need to be stored in the global memory.
Due to the limited bandwidth and high latency of global memory access, standard attention incurs significant memory I/O overhead when reading and writing $(\mathbf{S},\mathbf{P})$.

FlashAttention~\cite{dao2022flashattention} is an IO-aware technique designed to reduce memory overhead in attention operations.
It leverages online softmax to process the input matrices $\mathbf{Q}$, $\mathbf{K}$, and $\mathbf{V}$ in tiles, performing block-wise computation within fast shared memory.

\begin{algorithm*}[t]
\caption{Implementation of VecInfer.}
\label{alg: algorithm}

\begin{algorithmic}[1] 
\STATE \textbf{Input:} ${\mathbf{q}}\in \mathbb{R}^{1 \times D}$, ${\mathbf{K}}, \mathbf{V} \in \mathbb{R}^{N \times D}$, $\mathcal{C}_k ,\mathcal{C}_v \in \mathbb{R}^{2^b \times \frac{D}{M}}$, block size $B$. 
\STATE \textbf{Preprocessing:} $\tilde{\mathbf{q}}=\mathbf{q}\mathrm{diag}(\bm{\lambda})\mathbf{H}_D$, $\tilde{\mathbf{K}}=\mathbf{K}\mathrm{diag}(\bm{\lambda})^{-1}\mathbf{H}_D$. \hfill \textcolor{gray}{// \texttt{Dual transformation.}}
\STATE \textbf{Quantization:} $\tilde{\mathbf{K}}_{q}=\mathrm{VQ}(\tilde{\mathbf{K}},\mathcal{C}_k)$, ${\mathbf{V}}_{q}=\mathrm{VQ}({\mathbf{V}},\mathcal{C}_v)$. \hfill \textcolor{gray}{// \texttt{Vector quantization.}}
\STATE Compute $\tilde{\mathbf{q}}^{\prime}=\mathrm{reshape}(\tilde{\mathbf{q}},(M,\frac{D}{M}))$,  $\mathbf{lut} = \tilde{\mathbf{q}}^{\prime} \mathcal{C}_{k}^{\top}$.
\STATE Divide $\tilde{\mathbf{K}}_{q}, {\mathbf{V}}_{q}$ into $T=\left\lceil\frac{N}{B}\right\rceil$ blocks $\{\tilde{\mathbf{K}}_{q}^{(i)}\},  \{{\mathbf{V}}_{q}^{(i)}\}$.  \label{alg: divide}
\STATE Initialize $\mathbf{o} =(0)_{1 \times D} \in \mathbb{R}^{1 \times D}$, $\ell=(0),m= (-\infty)$ in SMEM.
\STATE Load $\mathcal{C}_{v}$ from GMEM to SMEM.
\STATE Load $\tilde{\mathbf{K}}_{q}^{(1)}$ from GMEM to SMEM.
\FOR{$i=1$ {\bfseries to} $T$} \label{alg: for}
\STATE Prefetch ${\mathbf{V}}_{q}^{(i)}$ from GMEM to SMEM.  \hfill \textcolor{gray}{//  \texttt{Asynchronous memory copy operation.}} \label{alg: load1}
\STATE Compute $\mathbf{s}^{(i)}=\mathrm{lookup}(\tilde{\mathbf{K}}_{q}^{(i)}, \mathbf{lut})/\sqrt{D}$. 
\STATE Compute $m^{new} = \mathrm{max}\{m, \mathrm{rowmax} (\mathbf{s}^{(i)})\}$, $\mathbf{p}^{(i)}=\mathrm{exp}(\mathbf{s}^{(i)}-m^{new})$.  \hfill \textcolor{gray}{// \texttt{Online softmax.}} 
\STATE Compute $\ell^{new}=\mathrm{exp}({m-m^{new}})\ell+\mathrm{rowsum}(\mathbf{p}^{(i)})$.
\STATE Wait for ${\mathbf{V}}_{q}^{(i)}$ to be loaded in SMEM.  \label{alg: wait}
\STATE Prefetch $\tilde{\mathbf{K}}_{q}^{(i+1)}$ from GMEM to SMEM.  \hfill \textcolor{gray}{//  \texttt{Asynchronous memory copy operation.}} \label{alg: load2}
\STATE Compute $\mathbf{o} = \mathrm{diag}(\mathrm{exp}({m-m^{new}}))\mathbf{o} + \mathbf{p}^{(i)} \mathrm{VQ}^{-1}(\mathbf{V}_{q}^{(i)}, \mathcal{C}_v)$.  \label{alg: pv}
\STATE Wait for $\tilde{\mathbf{K}}_{q}^{(i+1)}$ to be loaded in SMEM. \label{alg: wait2}
\STATE $\ell=\ell^{new}$, $m=m^{new}$.
\ENDFOR \label{alg: end_for}
\STATE Compute $\mathbf{o}=\mathrm{diag}(\ell)^{-1}\mathbf{o}$.
\STATE Compute $L=m+\mathrm{log}(\ell)$.
\STATE Write $\mathbf{o}$, $L$ to GMEM. 
\STATE Return the output $\mathbf{o}$ and the logsumexp $L$.
\end{algorithmic}
\end{algorithm*}

FlashAttention2~\cite{dao2024flashattention} improves GPU resource utilization by implementing optimized parallelization strategies. In contrast to FlashAttention, which performs parallelization across both the batch and head dimensions, FlashAttention2 focuses on parallelizing the query length dimension. Additionally, the computation is restructured by placing $\mathbf{Q}$ in the outer loop, while $\mathbf{K}$ and $\mathbf{V}$ are placed in the inner loop. 
FlashAttention-2 tiles $\mathbf{Q}\in \mathbb{R}^{N\times D}$ along the token dimension into blocks $\mathbf{Q}_i$ of size $B_q\times D$, resulting in $T_q=N/B_q$ blocks in total. Similarly, $\mathbf{K}, \mathbf{V}\in \mathbb{R}^{N\times D}$ are partitioned along the token dimension into blocks $\mathbf{K}_i$ and $\mathbf{V}_i$ of size $B_{kv}\times D$, yielding $T_{kv}=N/B_{kv}$ blocks.
The computation for each block is performed as follows:
\begin{equation}
\begin{aligned}
&\mathbf{S}_{ij}=\mathbf{Q}_i\mathbf{K}_j^\top / \sqrt{D},\\
&m_{ij}=\mathrm{max}\{m_{i,j-1},\mathrm{rowmax}(\mathbf{S}_{ij})\},\\
&\widetilde{\mathbf{P}}_{ij}=\mathrm{exp}(\mathbf{S}_{ij}-m_{ij}),\\
&\ell_{ij}=e^{m_{i,j-1}-m_{ij}}\ell_{i,j-1}+\mathrm{rowsum}(\widetilde{\mathbf{P}}_{ij}),\\
&\mathbf{O}_{ij}=\mathrm{diag}(e^{m_{i,j-1}-m_{ij}})\mathbf{O}_{i,j-1}+\widetilde{\mathbf{P}}_{ij}\mathbf{V}_j,
\end{aligned}
\end{equation}
where $(\mathbf{S}_{ij},\widetilde{\mathbf{P}}_{ij})\in \mathbb{R}^{B_q\times B_{kv}}$, $(m_{ij}, \ell_{ij}) \in \mathbb{R}^{B_q}$, and $\mathbf{O}_{ij} \in \mathbb{R}^{B_q \times D}$.
Finally, the output $\mathbf{O}_i$ is computed as follows:
\begin{equation}
\mathbf{O}_i=\mathrm{diag}(\ell_{i,T_{kv}})^{-1}\mathbf{O}_{i,T_{kv}}.
\end{equation}

FlashAttention3~\cite{shah2024flashattention3fastaccurateattention} further enhances attention computation by leveraging the architectural features of Hopper GPUs. FlashAttention3 employs three key optimizations.
First, it adopts producer–consumer warp specialization, assigning distinct warp groups to data loading and computation in order to hide transfer latency.
Second, it interleaves block-wise GEMMs with online softmax operations.
Third, it utilizes \texttt{FP8} hardware support to perform block quantization on the $\mathbf{Q}$, $\mathbf{K}$, and $\mathbf{V}$ matrices.

\subsection{Attention Varients}

LLM inference proceeds in two stages~\cite{zhangsurvey,gu-etal-2024-light,gu-etal-2025-adapt}: \textit{(i) the prefilling phase}, which processes the input prompt to produce the first output token; and \textit{(ii) the decoding phase}, which generates subsequent tokens autoregressively.

\begin{figure*}[t]
    \centering
    \begin{subfigure}[b]{\linewidth}
        \centering
        \includegraphics[width=\linewidth]{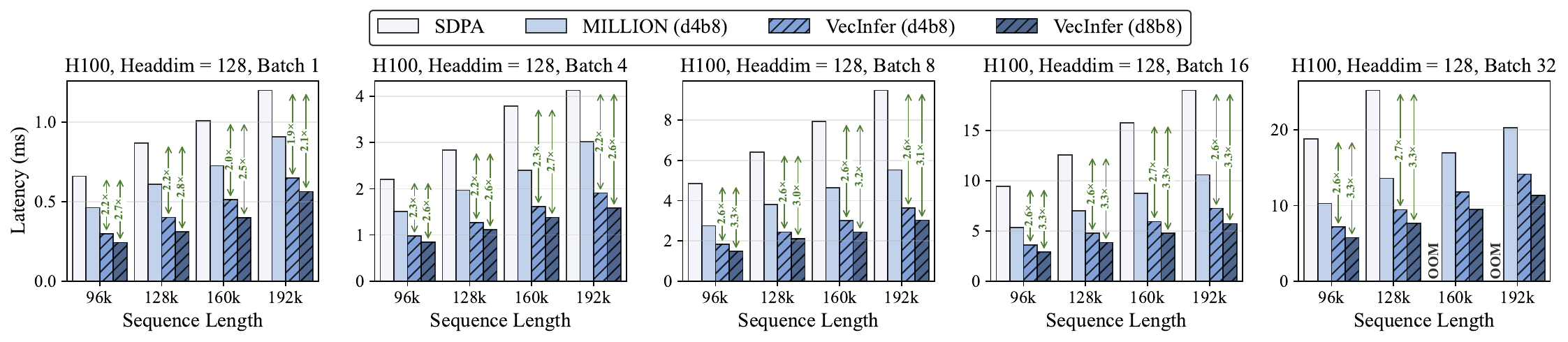}
        \caption{H100 80GB, headdim=128, query heads=32, KV heads=8}
    \end{subfigure}

    \begin{subfigure}[b]{\linewidth}
        \centering
        \includegraphics[width=\linewidth]{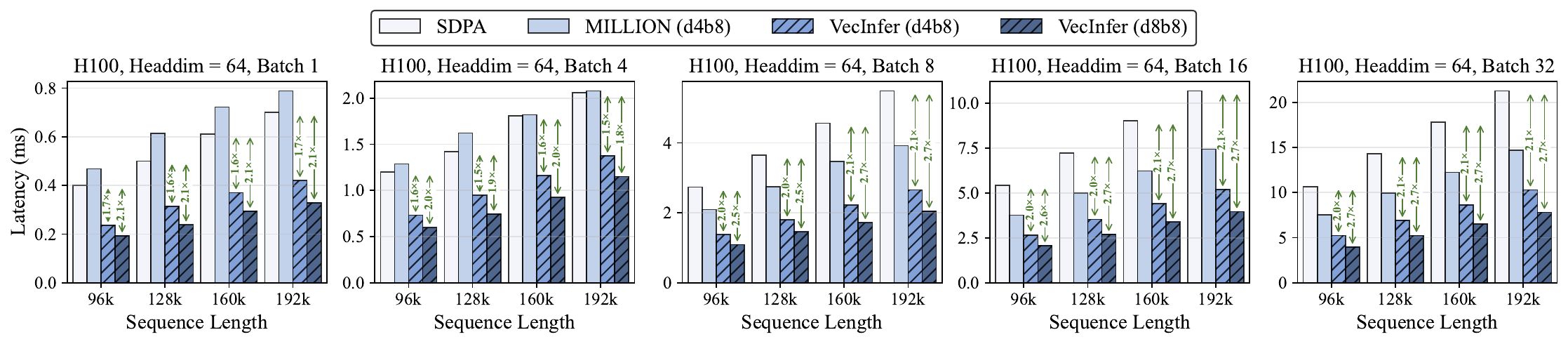}
        \caption{H100 80GB, headdim=64, query heads=32, KV heads=8}
    \end{subfigure}
    
    \begin{subfigure}[b]{\linewidth}
        \centering
        \includegraphics[width=\linewidth]{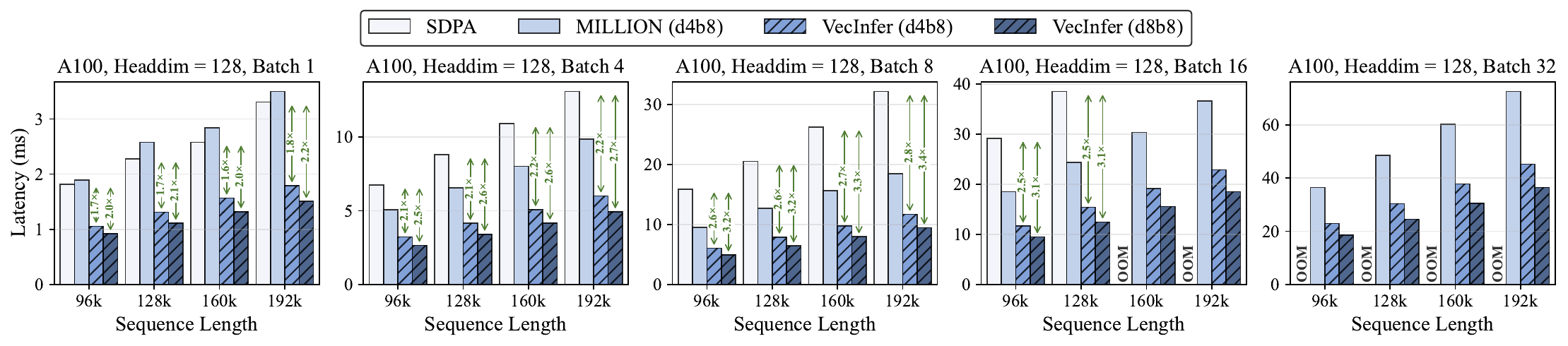}
        \caption{A100 40GB, headdim=128, query heads=32, KV heads=8}
    \end{subfigure}

    \begin{subfigure}[b]{\linewidth}
        \centering
        \includegraphics[width=\linewidth]{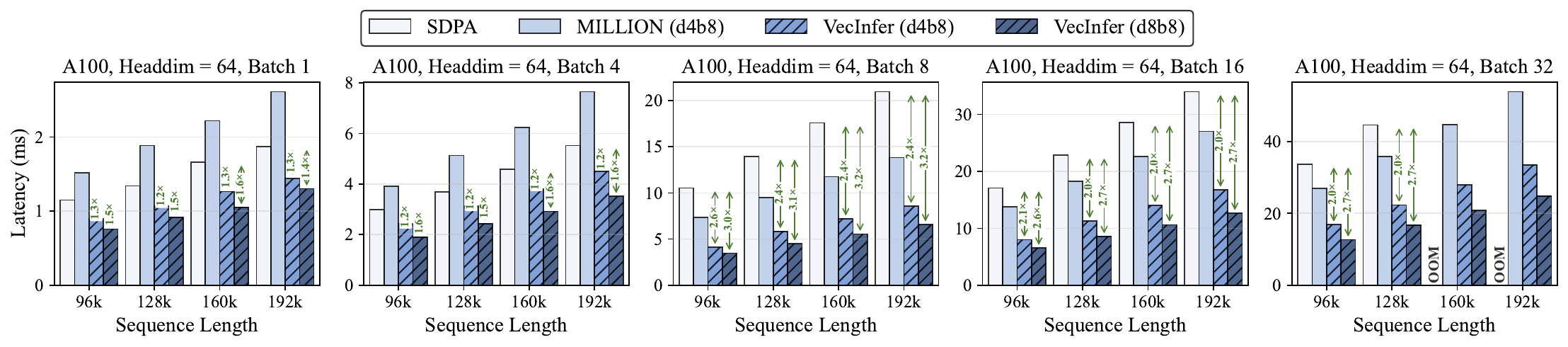}
        \caption{A100 40GB, headdim=64, query heads=32, KV heads=8}
    \end{subfigure}
    \caption{Comparison of kernel performance across different batch sizes and sequence lengths.}
    \label{fig: kernal_all}
\end{figure*}
For the compute-bound prefilling stage, existing methods focus on maximizing parallel processing capabilities and computational throughput.
Building on online softmax~\cite{milakov2018onlinenormalizercalculationsoftmax}, FlashAttention~\cite{dao2022flashattention,dao2024flashattention,shah2024flashattention3fastaccurateattention} proposes an IO-aware exact attention algorithm that uses tiling to minimize memory transfers between GPU high bandwidth memory and on-chip SRAM.
SageAttention~\cite{zhang2025sageattention,zhang2025sageattention2,zhang2025sageattention3microscalingfp4attention} is an efficient quantization method for attention that enhances the efficiency of attention computation while maintaining precision.
MInference~\cite{jiang2024minference} introduces hybrid sparse attention computation to accelerate LLM inference in the prefilling phase.
PAROAttention~\cite{zhao2025paroattentionpatternawarereorderingefficient} designs specialized sparsification and quantization techniques tailored to the unified block-wise pattern.
Recent studies~\cite{yuan-etal-2025-native,lu2025mobamixtureblockattention} have explored trainable sparse attention to accelerate attention computation.

\begin{figure*}[t]
  \centering
  \includegraphics[width=1\linewidth]{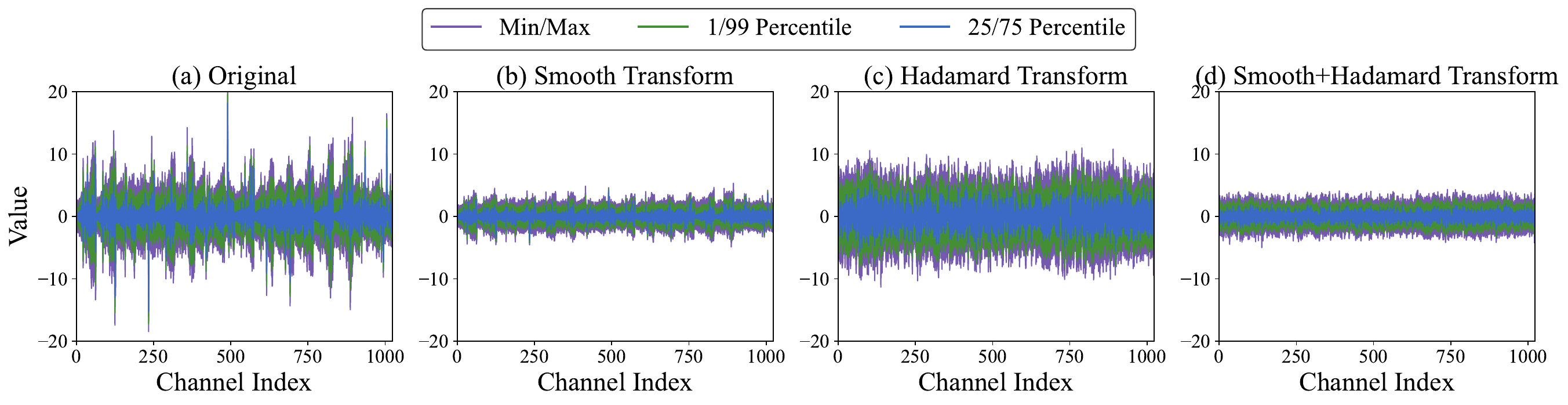} 
  \caption{Distribution of key cache for Llama-3.1-8B-Instruct (layer 16) under different transformations.}
  \label{fig: key_states_appendix}
\end{figure*}
\section{Implementation Details of Kernel}
\subsection{Kernel Fusion}
\label{appendix: kernel}

For the memory-bound decoding stage, existing methods focus on accelerating KV cache I/O operations.
Some methods~\cite{li2024snapkv,zhang2023h2o,liu2023scissorhands} are designed to reduce memory and computation costs through dynamic KV cache pruning. 
Quest~\cite{tang2024quest} and InfLLM~\cite{xiao2024infllm} retain the complete KV cache while retrieving only the most important tokens to reduce attention computation.
Quantization methods compress the KV cache into low-precision representations~\cite{liu2024kivi,hooper2024kvquant}, thereby reducing memory overhead.
FlashDecoding~\cite{hong2024flashdecodingfasterlargelanguage}, a variant of FlashAttention, is specifically designed to improve efficiency in long-context decoding.
To further reduce memory overhead and improve overall efficiency, BitDecoding~\cite{du2025bitdecodingunlockingtensorcores} enables efficient low-bit KV cache decoding by cooperatively leveraging CUDA Cores and Tensor Cores.

To reduce the overhead of reading/writing intermediate matrices in Equation~\eqref{eq: attention2}, we adopt fused dequantization-computation kernel, which improves both memory and latency efficiency.
Specifically, the implementation leverages fine-grained tiled computation and asynchronous pipeline execution, as outlined in Algorithm~\ref{alg: algorithm}.

Given the quantized codes $\tilde{\mathbf{K}}_{q},\mathbf{V}_{q} \in \mathbb{R}^{N \times M}$, a pre-computed lookup table $\mathbf{lut} \in \mathbb{R}^{M \times 2^b}$, and a value codebook $\mathcal{C}_v \in \mathbb{R}^{2^b \times \frac{D}{M}}$, we aim to compute the attention output $\mathbf{o} \in \mathbb{R}^{1 \times D}$ and write it back to global memory.
The algorithm begins by partitioning $\tilde{\mathbf{K}}_{q}$ and ${\mathbf{V}}_{q}$ into $T=\left\lceil\frac{N}{B}\right\rceil$ blocks, each of size $B \times M$ (line~\ref{alg: divide}).
Attention is then computed in a block-wise manner, with iterations described in lines~\ref{alg: for}-\ref{alg: end_for}.
To minimize memory I/O, we adopt the online softmax scheme~\cite{dao2022flashattention}, which incrementally rescales partial results from each block to ensure a correct final output.

\begin{figure}[t]
  \centering
  \includegraphics[width=1\linewidth]{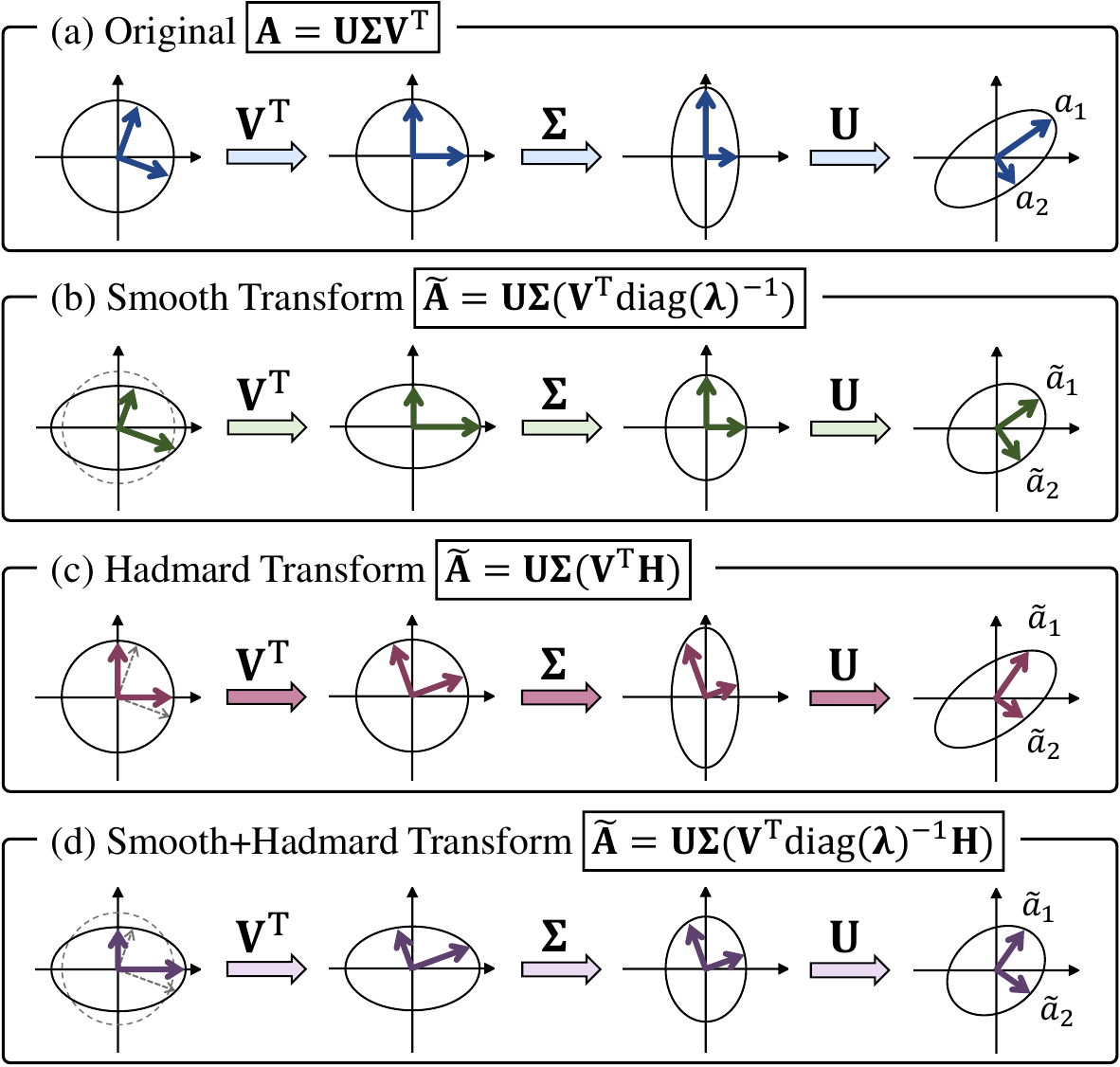} 
  \caption{Transformation from $\mathbf{V}^\top$ to $\mathbf{A}$ via SVD.}
  \label{fig: svd_appendix}
\end{figure}

To fully utilize CUDA cores, we leverage the \texttt{memcpy\_async} API to overlap memory transfers with computation.
Specifically, while computing $\mathbf{s}^{(i)}$, we asynchronously load $\mathbf{V}_q^{(i)}$ from global memory into shared memory (line~\ref{alg: load1}), thereby hiding memory latency.
The first synchronization (line~\ref{alg: wait}) ensures that this transfer has completed before the values are used.
Similarly, during the computation of $\mathbf{o}^{(i)}$, we prefetch $\tilde{\mathbf{K}}_q^{(i+1)}$ for the next iteration (line~\ref{alg: load2}).
The second synchronization (line~\ref{alg: wait2}) then guarantees that the prefetched data are ready at the beginning of iteration $i{+}1$.

\subsection{Kernel Latency Comparison}

Figure~\ref{fig: kernal_all} compares the speed of the VecInfer kernel against baselines using headdim=64 and headdim=128, with sequence lengths ranging from 96k to 192k, on A100 (40GB) and H100 (80GB) GPUs.

\section{Details of Different Transformations}
\label{appendix: transformations}
\begin{table*}[t]
\setlength{\tabcolsep}{2.2pt}
\renewcommand{\arraystretch}{0.97}
\centering
\resizebox{\textwidth}{!}{
\begin{tabular}{l|cc|ccccccccccccc|c}
\toprule

\multirow{3}{*}{\textbf{Avg. bit}} &\multirow{3}{*}{\textbf{Key Config}}  &\multirow{3}{*}{\textbf{Value Config}} & \multicolumn{2}{c}{\textbf{SD.QA}} & \multicolumn{2}{c}{\textbf{MD.QA}}& \multicolumn{2}{c}{\textbf{Sum.}}& \multicolumn{3}{c}{\textbf{FS.L}}& \multicolumn{2}{c}{\textbf{Code}} & \multicolumn{2}{c}{\textbf{Synth.}} &\multirow{3}{*}{\textbf{Avg.}} \\

\cmidrule(lr){4-5}\cmidrule(lr){6-7}\cmidrule(lr){8-9}\cmidrule(lr){10-12}\cmidrule(lr){13-14}\cmidrule(lr){15-16}
&&& Qspr & MulF & HQA & WMQA & GRpt & MulN & TREC & SMSM & TriQA & Repo & LCC & PsgC & PsgR \\

\midrule

\multirow{3}{*}{1.5} & 1.5-bit (\texttt{d8b12}) & 1.5-bit (\texttt{d8b12}) & 43.7 & 52.2 & 54.7 & 46.2  & 29.2 & 26.1 & 71.0 & 39.4 & 91.7 & 51.4 & 60.2 & 7.6 & 99.5 & 51.8 \\
 & 1-bit (\texttt{d8b8}) & 2-bit (\texttt{d4b8}) & 37.2 & 45.4 & 52.1 & 41.9 & 21.7 & 23.8 & 64.5 & 36.9 & 88.4 & 43.9 & 52.6 & 8.0 & 96.0 & 47.1 \\
 &  2-bit (\texttt{d4b8}) & 1-bit (\texttt{d8b8}) & 43.5 & 52.5 & 54.7 & 46.0 & 27.2 & 25.4 & 72.5 & 39.2 & 92.3 & 52.6 & 60.5 & 7.9 & 99.5 & 51.9 \\
\midrule
\multirow{3}{*}{1.25} & 1.25-bit (\texttt{d8b10}) & 1.25-bit (\texttt{d8b10}) & 38.2 & 47.4 & 54.1 & 44.4 & 25.4 & 24.3 & 66.0 & 38.2 & 90.5 & 48.1 & 56.4 & 7.7 & 97.5 & 49.1 \\
 & 1-bit (\texttt{d8b8}) & 1.5-bit (\texttt{d8b12}) & 36.3 & 44.9 & 52.4 & 42.2 & 21.1 & 23.2 & 62.5 & 36.0 & 87.8 & 43.2 & 52.9 & 8.0 & 97.0 & 46.7 \\
 & 1.5-bit (\texttt{d8b12}) & 1-bit (\texttt{d8b8}) & 41.2 & 49.6 & 54.2 & 45.4 & 25.7 & 24.2 & 69.0 & 39.1 & 90.8 & 50.4 & 57.7 & 7.2 & 99.5 & 50.3\\

\bottomrule
 
\end{tabular}}

\caption{Accuracy comparison under different quantization bit-width configurations using Llama-3.1-8B-Instruct. }
\label{tab: sensitivity}
\end{table*}

\begin{table*}[t]
\setlength{\tabcolsep}{4.5pt}
\renewcommand{\arraystretch}{0.94}
\centering
\resizebox{\textwidth}{!}{
\begin{tabular}{l|ccccccccccccc|c}
\toprule

\multirow{3}{*}{\textbf{Dataset}} & \multicolumn{2}{c}{\textbf{SD.QA}} & \multicolumn{2}{c}{\textbf{MD.QA}}& \multicolumn{2}{c}{\textbf{Sum.}}& \multicolumn{3}{c}{\textbf{FS.L}}& \multicolumn{2}{c}{\textbf{Code}} & \multicolumn{2}{c}{\textbf{Synth.}} &\multirow{3}{*}{\textbf{Avg.}} \\

\cmidrule(lr){2-3}\cmidrule(lr){4-5}\cmidrule(lr){6-7}\cmidrule(lr){8-10}\cmidrule(lr){11-12}\cmidrule(lr){13-14}
& Qspr & MulF & HQA & WMQA & GRpt & MulN & TREC & SMSM & TriQA & Repo & LCC & PsgC & PsgR \\
\midrule
Qspr & 46.1 & 53.1 & 55.0 & 46.5 & 31.2 & 26.5 & 72.0   & 41.6 & 92.2 & 53.2 & 60.9 & 7.9 & 98.5 & 52.7  \\
MulF & 46.2 & 52.5 & 54.8 & 46.3 & 31.3 & 26.3 & 72.5 & 42.8 & 92.4 & 53.8 & 60.6 & 7.3 & 99.5 & 52.8 \\
HQA & 44.2 & 53.1 & 55.1 & 47.0 & 32.1 & 26.8 & 72.5 & 42.3 & 91.9 & 55.0 & 61.1 & 7.6 & 99.5 & 53.0 \\
WMQA & 45.7 & 52.0 & 55.3 & 46.5 & 31.2 &  26.4  & 72.5 & 41.5 & 92.4 & 54.6 & 61.1 & 7.7 & 99.0 &  52.8  \\
TriQA & 43.8 & 52.2 & 55.6 & 47.3 & 30.7 &  26.7  & 72.0 & 41.8 & 91.7 & 53.7 & 61.4 & 7.5 & 100.0 &  52.7  \\

\bottomrule
 
\end{tabular}}
\caption{Evaluation results using codebooks clustered from different datasets under 2-bit quantization with Llama-3.1-8B-Instruct.}
\label{tab: codebook}
\end{table*}
\begin{table*}[p]
\setlength{\tabcolsep}{2.2pt}
\renewcommand{\arraystretch}{1}
\centering
\resizebox{\textwidth}{!}{
\begin{tabular}{lcc|ccccccccccccc|c}
\toprule

\multirow{3}{*}{\textbf{Method}} &\multirow{3}{*}{\textbf{Avg. bit}}  &\multirow{3}{*}{\textbf{Config}} & \multicolumn{2}{c}{\textbf{SD.QA}} & \multicolumn{2}{c}{\textbf{MD.QA}}& \multicolumn{2}{c}{\textbf{Sum.}}& \multicolumn{3}{c}{\textbf{FS.L}}& \multicolumn{2}{c}{\textbf{Code}} & \multicolumn{2}{c}{\textbf{Synth.}} &\multirow{3}{*}{\textbf{Avg.}} \\

\cmidrule(lr){4-5}\cmidrule(lr){6-7}\cmidrule(lr){8-9}\cmidrule(lr){10-12}\cmidrule(lr){13-14}\cmidrule(lr){15-16}
&&& Qspr & MulF & HQA & WMQA & GRpt & MulN & TREC & SMSM & TriQA & Repo & LCC & PsgC & PsgR \\

\midrule

\rowcolor[rgb]{0.957,0.957,0.957}
\textit{Llama-3.1-8B} & 16 & - & 45.9 & 53.8 & 55.2 & 46.6 & 34.6 & 27.5 & 72.5 & 43.8 & 91.6 & 56.4 & 63.2 & 8.0 & 99.5 & 53.7  \\
\midrule

KIVI & 4.25 & \texttt{b4g128} & 45.6 & 53.8 & 54.9 & 47.1 & 34.6 & 27.1 & 72.5 & 44.0 & 92.4 & 56.7 & 63.2 & 7.6 & 99.5 &53.6\\

MILLION & 4 & \texttt{d2b8} & 46.5 & 53.5 & 55.6 & 46.1 & 34.1 & 26.9 & 72.5 & 42.6 & 92.1 & 55.1 & 62.9 & 8.0 & 99.5 & 53.5\\

\rowcolor[rgb]{0.87,0.94,1}
VecInfer & 4 & \texttt{d2b8} & 45.6 & 54.0 & 55.3 & 46.6   & 34.3 & 26.8 & 72.5 & 43.1  & 92.0 & 56.0 & 63.1 & 7.6 & 99.5  & \textbf{53.6} \\

\midrule

\rowcolor[rgb]{0.957,0.957,0.957}
\textit{Mistral-7B} & 16 & - & 38.6 & 49.7 & 51.0 & 36.4 & 34.3 & 26.5 & 76.0 & 47.6 & 88.5 & 60.6 & 59.4 & 6.0 & 96.5 & 51.5\\
\midrule

KIVI & 4.25 & \texttt{b4g128} & 38.0 & 50.4 & 51.1 & 36.6 & 34.4 & 26.6 & 76.0 & 46.9 & 88.9 & 58.8 & 58.3 & 5.0 & 96.0 & 51.3\\

MILLION & 4 & \texttt{d2b8} & 36.9 & 51.2 & 51.0 & 36.0 & 33.6 & 26.4 & 75.5 & 45.8 & 88.9 & 59.5 & 58.6 & 2.5 & 93.0 & 50.7\\

\rowcolor[rgb]{0.87,0.94,1}
VecInfer & 4 & \texttt{d2b8} & 38.4 & 49.8 & 51.1 & 35.9 & 34.1 & 24.0 & 76.0   & 47.4 & 88.7 & 60.7 & 59.3 & 6.0    & 96.0  & \textbf{51.4}\\
\midrule

\rowcolor[rgb]{0.957,0.957,0.957}
\textit{Qwen2.5-14B} & 16 & - & 45.4 & 53.9 & 61.6 & 57.9 & 29.7 & 21.9 & 76.5 & 47.7 & 90.1 & 48.8 & 61.3 & 9.0 & 98.6 &54.0 \\

\midrule
KIVI & 4.25 & \texttt{b4g128} & 45.6 & 52.2 & 61.5 & 58.5 & 29.3 & 21.8 & 77.0 & 47.5 & 90.4 & 49.2 & 61.6 & 10.9 & 98.2 & 54.1\\

MILLION & 4 & \texttt{d2b8} & 43.5 & 51.7 & 61.0 & 57.7 & 28.0 & 22.6 & 76.5 & 46.4 & 89.9 & 49.0 & 60.2 & 10.9 & 93.5 & 53.1\\

\rowcolor[rgb]{0.87,0.94,1}
VecInfer & 4 & \texttt{d2b8} & 45.8 & 54.0 & 61.9 & 58.3 & 29.5 & 21.5 & 77.0   & 47.7 & 90.0 & 49.2 & 61.3  & 10.0 & 98.9 & \textbf{54.3} \\

\bottomrule
 
\end{tabular}}

\caption{Additional LongBench evaluation results under 4-bit quantization.}
\label{tab: longbench2}
\end{table*}

To analyze the effects of different transformations, we use SVD, which factorizes a matrix as $\mathbf{A}=\mathbf{U\Sigma V}^\top$, where $\mathbf{U}$ and $\mathbf{V}$ are orthogonal and $\mathbf{\Sigma}$ is diagonal.
SVD decomposes transformations into rotation ($\mathbf{U}$, $\mathbf{V}$) and stretch ($\mathbf{\Sigma}$) components~\cite{lee2024infinigen}.
For example, Figure~\ref{fig: svd_appendix}(a) shows how the column vectors of $\mathbf{V}^\top$ are rotated and stretched to form the column vectors $a_1$ and $a_2$ of $\mathbf{A}$, which represent the maximum and minimum values, respectively.
Figure~\ref{fig: svd_appendix}(b) and Figure~\ref{fig: svd_appendix}(c) demonstrate that smooth and Hadamard transformations, respectively, reduce the magnitude difference between vectors $\tilde{a}_1$ and $\tilde{a}_2$ by stretching and rotating the column vectors of $\mathbf{V}^\top$.
In Figure~\ref{fig: svd_appendix}(d), combining these two transformations achieves optimal balance between vectors $\tilde{a}_1$ and $\tilde{a}_2$.
Overall, Figure~\ref{fig: key_states_appendix} further demonstrates that applying these transformations individually yields suboptimal uniformity, whereas their combination achieves the most uniform distribution.

\section{Quantization Sensitivity of Key/Value Cache}
\label{appendix: sensitivity}

Existing research~\cite{dong2024qaqqualityadaptivequantization} has demonstrated that the presence of outliers amplifies quantization errors in key cache, leading to distinct quantization sensitivities between key cache and value cache. 
However, our findings reveal that even with outlier suppression through transformation, the transformed key cache still exhibits higher quantization sensitivity than value cache, as illustrated in Table~\ref{tab: sensitivity}.
Our strategy involves implementing mixed precision by assigning separate storage precisions for key cache and value cache.

\section{Additional Experiments}
\subsection{Ablation Study: Task-independent Codebook}
\label{appendix: codebook}
As shown in Table~\ref{tab: codebook}, the performance remains nearly consistent when using codebooks pre-trained on different datasets, suggesting that the learned codebooks generalize well and are effectively task-independent.

\subsection{Accuracy Evaluation}

Table~\ref{tab: longbench2} reports additional results on LongBench. The results demonstrate that all models tested with VecInfer maintain lossless performance under 4-bit quantization, and VecInfer consistently outperforms other methods.

\section{Additional Details of Benchmarks}
\label{appendix: benchmark}

\textbf{LongBench}: A comprehensive benchmark covering six categories of evaluation tasks—single/multi-document question answering, summarization, code completion, synthetic tasks, and few-shot learning. Detailed information on all 13 datasets included in LongBench is presented in Table~\ref{tab: longbench_detail}.

\noindent \textbf{GSM8K}: A carefully selected dataset comprising 1,319 grade-school math problems.

\noindent \textbf{MATH500}: A challenging math dataset comprising 500 problems from high school math competitions.

\noindent \textbf{AIME24}: A collection of 30 challenging mathematical problems sourced from the 2024 American Invitational Mathematics Examination.

\noindent \textbf{AMC2023}: A set of 40 problems from the 2023 American Mathematics Competitions, designed to rigorously test reasoning and problem-solving skills.

\begin{table*}[p]
\small
\centering
\begin{tabular}{lccccc}
\toprule
\textbf{Label} & \textbf{Task} & \textbf{Capability} & \textbf{Metric} & \textbf{Avg. len}  & \textbf{\#data}\\
\midrule
Qspr & Qasper & Single-Doc. QA (SD.QA) & F1 & 3,619  & 200  \\
MulFi & MultiFieldQA-en & Single-Doc. QA (SD.QA) & F1 & 4,559  & 150  \\
HQA & HotpotQA & Multi-Doc. QA (MD.QA) & F1 & 9,151  & 200  \\
WMQA & 2WikiMultihopQA & Multi-Doc. QA (MD.QA)& F1 & 4,887  & 200\\
GRpt & GovReport & Summarization (Sum.)& Rouge-L & 8,734  & 200  \\
MulN & MultiNews & Summarization (Sum.)& Rouge-L & 2,113  & 200 \\
TREC  & TREC & Few-shot Learning (FS.L)& Accuracy (CLS) & 5,177  & 200\\
SMSM & SAMSum & Few-shot Learning (FS.L)& Rouge-L & 6,258  & 200 \\
TriQA & TriviaQA & Few-shot Learning (FS.L)& F1 & 8,209  & 200 \\
Lcc & LCC & Code Completion (Code) & Edit Sim & 1,235   & 500\\
Repo & RepoBench-P & Code Completion (Code) & Edit Sim & 4,206  & 500  \\ 
PsgC & PassageCount & Synthetic (Synth.)& Accuracy (EM) & 11,141  & 200 \\
PsgR & PassageRetrieval-en & Synthetic (Synth.) & Accuracy (EM) & 9,289  & 200  \\
\bottomrule
\end{tabular}%
 \caption{Details of LongBench.}
 \label{tab: longbench_detail}
\end{table*}

\begin{table*}[p]
\small
\centering
\begin{tabular}{lcccc}
\toprule
\textbf{Dataset} & \textbf{Max Output} &\textbf{Responses} & \textbf{Avg. len} & \textbf{\#data}\\
\midrule

GSM8K &16,384 & 1 & 111 & 1,319\\
MATH500 & 16,384 & 1 & 121 & 500\\
AIME2024 & 32,768 & 8 & 156 & 30\\
AMC2023 & 16,384 & 8 & 137 & 40\\

\bottomrule
\end{tabular}%
 \caption{Details of mathematical reasoning benchmarks.}
 \label{tab: math_detail}
\end{table*}

\section{Information About Use Of Ai Assistants}
In this paper, we only use AI tools for grammar checking and code completion.

\newpage
\section{Proof of Lemma}

\begin{defbox}
\begin{customlemma}{1}[Hadamard]\label{lemma: Hadamard} 
For key states $\mathbf{K} \!\in \!\mathbb{R}^{N \times D}$ with $\mathrm{sign}(K_{i,j}) \overset{\text{i.i.d.}}{\sim} \mathrm{Uniform}\{-1, +1\}$, and a Hadamard matrix $\mathbf{H} \in \mathbb{R}^{D \times D}$ constructed as in Equation~\eqref{eq: hadamard}, the transformed matrix $\tilde{\mathbf{K}}=\mathbf{KH}$ exhibits approximately Gaussian distribution by the central limit theorem, thereby redistributing the outliers of $\mathbf{K}$.
\end{customlemma}
\end{defbox}

\begin{proof}
We denote the $(i,j)$-th entry of matrices $\mathbf{K}$, $\mathbf{H}$, and $\tilde{\mathbf{K}}$ as $K_{i,j}$, $H_{i,j}$, and $\tilde{K}_{i,j}$, respectively.
Consider any element of $\tilde{\mathbf{K}}$:
\begin{equation}
\tilde{K}_{i,j} = \sum_{l=1}^{D} K_{i,l} H_{l,j} 
= \sum_{l=1}^{D} |K_{i,l}| \cdot \epsilon_{i,l} \cdot H_{l,j},
\end{equation}
where $\epsilon_{i,l} = \mathrm{sign}(K_{i,l})\overset{\text{i.i.d.}}{\sim} \mathrm{Uniform}\{-1, +1\}$, and $H_{l,j} \in \left\{-\frac{1}{\sqrt{D}}, +\frac{1}{\sqrt{D}}\right\}$.
Then:
\begin{equation}
\epsilon_{i,l} \cdot H_{l,j} \overset{\text{i.i.d.}}{\sim} \mathrm{Uniform}\left\{-\frac{1}{\sqrt{D}}, +\frac{1}{\sqrt{D}}\right\}.
\end{equation}
The expectation of \(\tilde{K}_{i,j}\) is given by:
\begin{equation}
\begin{split}
\mathbb{E}[\tilde{K}_{i,j}] &= \mathbb{E}\left[\sum_{l=1}^{D} |K_{i,l}| \cdot \epsilon_{i,l} \cdot H_{l,j}\right]\\ &= \sum_{l=1}^{D} |K_{i,l}| \cdot \mathbb{E}[\epsilon_{i,l} \cdot H_{l,j}] = 0,
\end{split}
\end{equation}
The variance of \(\tilde{K}_{i,j}\) is computed as follows:
\begin{equation}
\begin{split}
\mathrm{Var}(\tilde{K}_{i,j}) &= \mathrm{Var}\left(\sum_{l=1}^{D} |K_{i,l}| \cdot \epsilon_{i,l} \cdot H_{l,j}\right)  \\ &= \sum_{l=1}^{D} |K_{i,l}|^2 \cdot \mathrm{Var}(\epsilon_{i,l} \cdot H_{l,j}).
\end{split}
\end{equation}
As $\mathbb{E}[\epsilon_{i,l} \cdot H_{l,j}] = 0$, we have:
\begin{equation}
\mathrm{Var}(\epsilon_{i,l} \cdot H_{l,j}) = \mathbb{E}[(\epsilon_{i,l} \cdot H_{l,j})^2] = \frac{1}{D}.
\end{equation}
Thus, the variance of \(\tilde{K}_{i,j}\) becomes:
\begin{equation}
\mathrm{Var}(\tilde{K}_{i,j}) = \sum_{l=1}^{D} |K_{i,l}|^2 \cdot \frac{1}{D}.
\end{equation}
By the Lindeberg-Feller Central Limit Theorem, as \(D \to \infty\), the sum \(\tilde{K}_{i,j}\) converges in distribution to a Gaussian random variable:
\begin{equation}
\tilde{K}_{i,j} \overset{d}{\to} \mathcal{N}\left(0, \frac{1}{D}\sum_{l=1}^{D} K_{i,l}^2\right).
\end{equation}
\end{proof}

\end{document}